\documentclass[10pt,journal,compsoc]{IEEEtran}
 \usepackage{mathrsfs}
 \usepackage{bm}
 \usepackage{amsfonts}
\usepackage{CJK}
\usepackage{amsmath}
\usepackage{algorithm}
\usepackage{algorithmic}
\usepackage{bm}
\usepackage{indentfirst}
\usepackage{multirow}
\usepackage{graphicx}
\usepackage{subfigure}
 \usepackage{verbatim}
 \usepackage{url}

\usepackage{array}
\newcolumntype{I}{!{\vrule width 3pt}}
\newlength\savewidth
\newcommand\shline{\noalign{\global\savewidth\arrayrulewidth
                            \global\arrayrulewidth 1.5pt}%
                   \hline
                   \noalign{\global\arrayrulewidth\savewidth}}

\ifCLASSOPTIONcompsoc
  \usepackage[nocompress]{cite}
\else
  \usepackage{cite}

\fi
\ifCLASSINFOpdf
\else
\fi
\begin{document}
%
\title{Robust Online Matrix Factorization for \\Dynamic Background Subtraction}
%
%
%
%

\author{Hongwei Yong,~\IEEEmembership{}
        Deyu Meng,~
        Wangmeng Zuo,~
        Lei Zhang
\IEEEcompsocitemizethanks{\IEEEcompsocthanksitem Hongwei Yong and Deyu Meng (corresponding author) are with School of Mathematics and Statistics and Ministry of Education Key Lab of Intelligent Networks and Network Security, Xi'an Jiaotong University, Shaanxi, P.R. China.
(e-mail:dymeng@mail.xjtu.edu.cn).\protect
\IEEEcompsocthanksitem Wangmeng Zuo is with School of Computer Science and Technology, Harbin Institute of Technology, Harbin P.R. China.
\IEEEcompsocthanksitem Lei Zhang is with the Biometrics Research Center, Department of Computing, the Hong Kong Polytechnic University, Kowloon, Hong Kong.}
}

\IEEEtitleabstractindextext{%
\begin{abstract}We propose an effective online background subtraction method, which can be robustly applied to practical videos that have variations in both foreground and background. Different from previous methods which often model the foreground as Gaussian or Laplacian distributions, we model the foreground for each frame with a specific mixture of Gaussians (MoG) distribution, which is updated online frame by frame.
Particularly, our MoG model in each frame is regularized by the learned foreground/background knowledge in previous frames. This makes our online MoG model highly robust, stable and adaptive to practical foreground and background variations.
The proposed model can be formulated as a concise probabilistic MAP model, which can be readily solved by EM algorithm. We further embed an affine transformation operator into the proposed model, which can be automatically adjusted to fit a wide range of video background transformations and make the method more robust to camera movements. With using the sub-sampling technique, the proposed method can be accelerated to execute more than 250 frames per second on average, meeting the requirement of real-time background subtraction for practical video processing tasks. The superiority of the proposed method is substantiated by extensive experiments implemented on synthetic and real videos, as compared with state-of-the-art online and offline background subtraction methods.
\end{abstract}

}

\maketitle

\IEEEdisplaynontitleabstractindextext

%
\IEEEpeerreviewmaketitle

\begin{figure*}[!]
\centering
\subfigure[$Bootstrap$ sequence]{
\label{Fig.sub.1}
\includegraphics[width=0.38\textwidth]{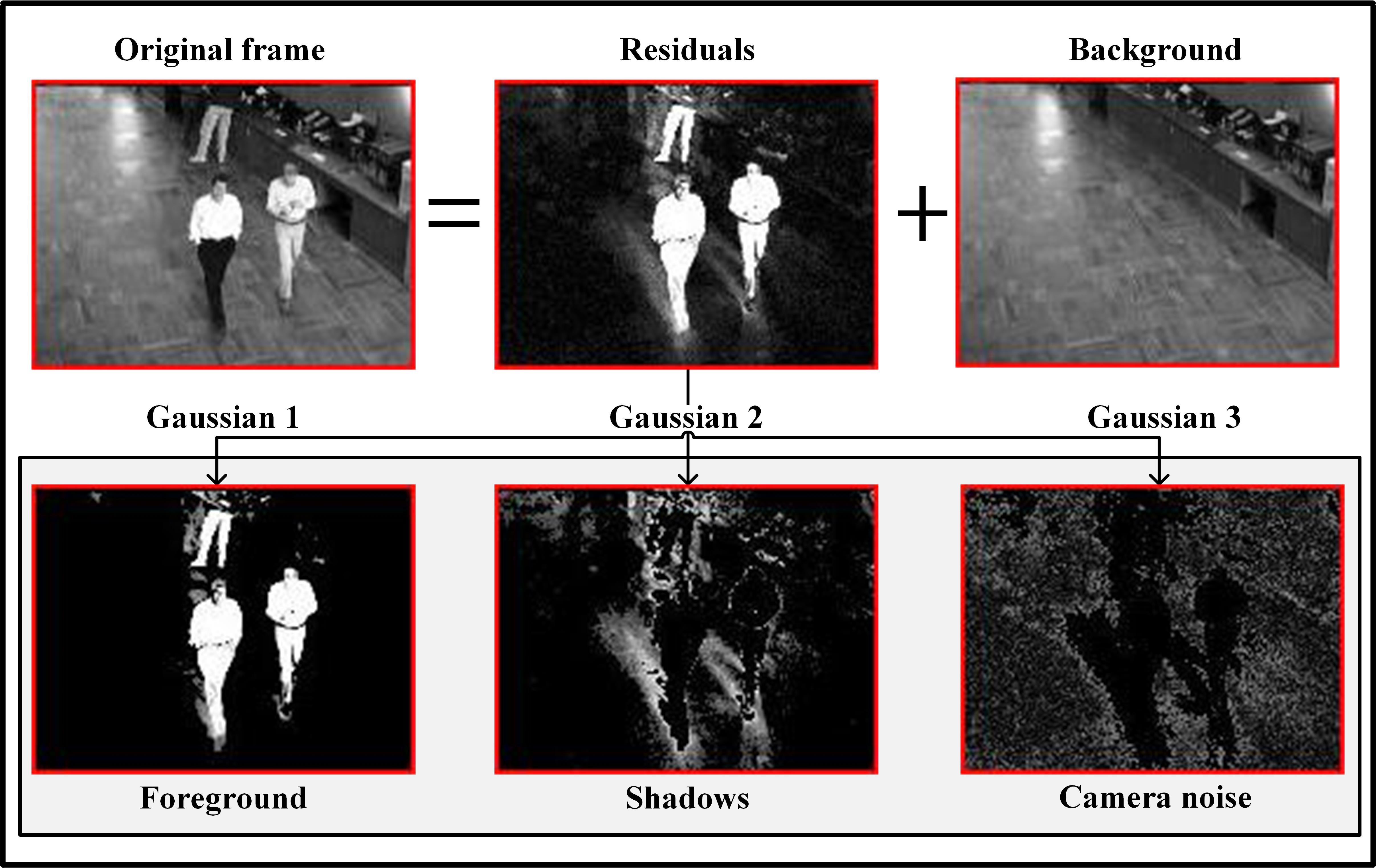}}
\subfigure[$Campus$ sequence]{
\label{Fig.sub.2}
\includegraphics[width=0.38\textwidth]{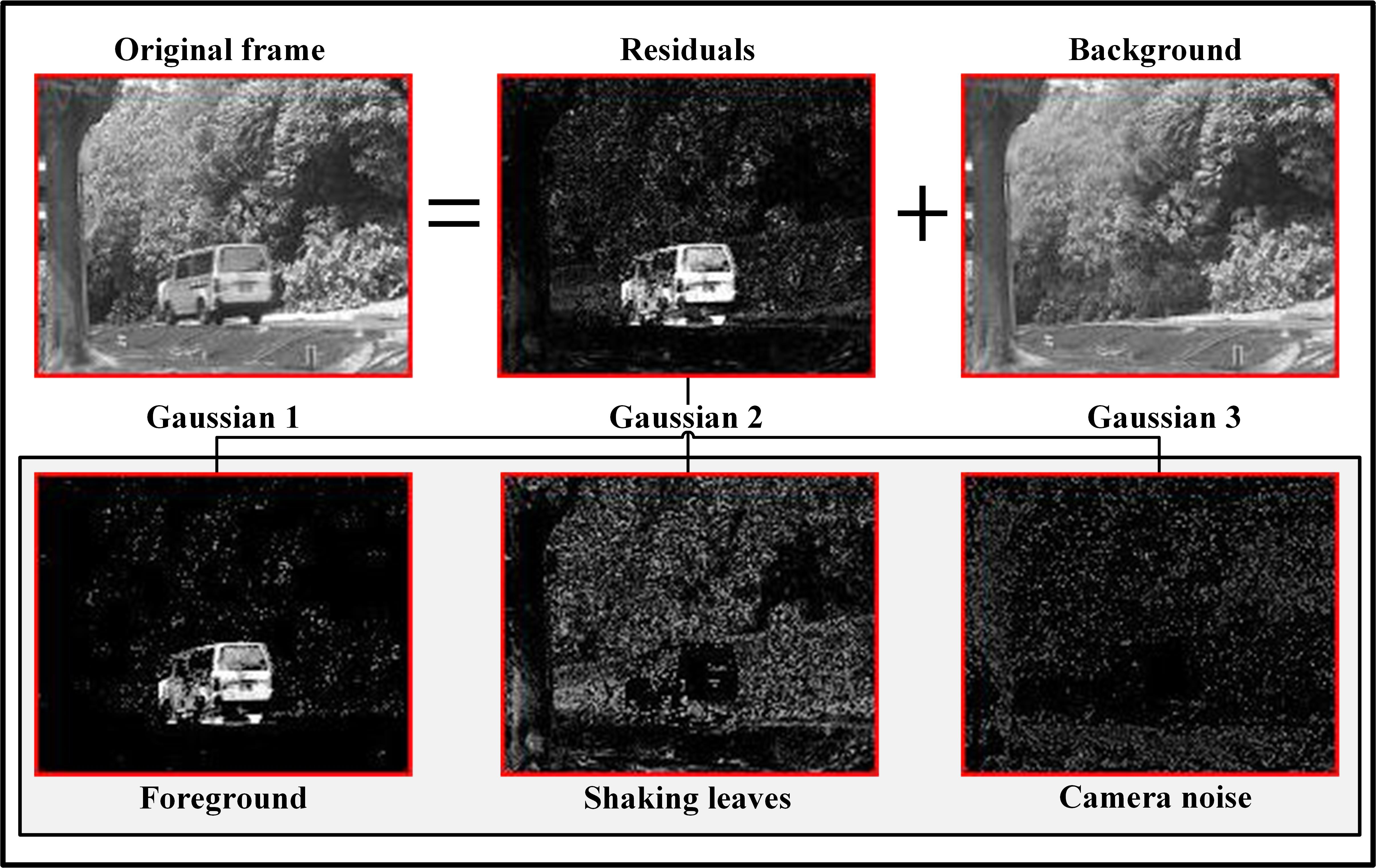}}
\vspace{-4mm}
\caption{Background subtraction results by the proposed OMoGMF method on ($a$) $bootstrap$ sequence; ($b$) $campus$ sequence. First row (from left to right): Original frame, noise and background extracted by the proposed method. Second row: Three noise components (after scale processing) extracted by the method, corresponding to the moving object, the shadow along the object, weak camera noise (for (a)) and the moving object, leaves shaking variance, weak camera noise (for (b)), respectively.
}
\label{figure:f1}
\vspace{-4mm}
\end{figure*}

\IEEEraisesectionheading{\section{Introduction}\label{sec:introduction}}
\IEEEPARstart{V}{ideo} processing is one of the main branches in image processing and computer vision, which is targeted to extract knowledge from videos collected from real scenes. As an essential and fundamental research topic in video processing, background subtraction
has been attracting increasing attention in the recent years. The main aim of background subtraction is to separate moving object foreground from the background in a video, which always makes the subsequent video processing tasks easier and more efficient. Typical applications of background subtraction include object tracking~\cite{beleznai2006multiple}, urban traffic detection~\cite{cheung2005robust}, long-term scene monitoring~\cite{senior2011interactive}, 
video compression~\cite{cao2015novel} and so on.

The initial strategies proposed to handle background subtraction are to directly distinguish background pixels from foreground ones through some simple statistical measures, like the median(mean) model~\cite{lee2002background,mcfarlane1995segmentation} and some histogram models~\cite{zheng2006extracting}. Later, more elaborate statistical models, like the MOG~\cite{stauffer1999adaptive} and MOGG~\cite{allili2007robust} models, were presented to better deliver the distributions of the image pixels located in the background. These methods, however,
 ignore very useful video structure knowledge, like temporal similarity of background scene and spatial contiguity of foreground objects, and thus always cannot guarantee a good performance especially under complex scenarios. In recent decades, low-rank subspace learning models~\cite{bouwmans2014robust,bouwmans2016decomposition} represent a new trend and achieve state-of-the-art performance for this task due to their better consideration of video structure knowledge both in foreground and background. Especially, these methods assume a rational low-rank structure for video backgrounds, which encodes the similarity of video backgrounds along time, and mostly consider useful prior foreground structures, like sparsity and spatial continuity. Some typical models along this line are
~\cite{candes2011robust},~\cite{wang2012probabilistic}, ~\cite{zhou2013moving},~\cite{zhou2011godec}, ~\cite{meng2013robust}.

Albeit substantiated to be effective in some video sequences with fixed lengthes, there is still a gap of utilizing such offline methodologies to real video processing applications. Specifically, it is known that the amount of videos nowadays is dramatically increasing from surveillance cameras scattered all over the world. This not only makes it critical to calculate background subtraction from such large \mbox{amount} of videos, but also urgently requires to construct real-time techniques to handle the constantly emerging videos. Online subspace learning has thus become an important issue to alleviate this efficiency issue. Very recently, multiple online methods for background subtraction have been designed \cite{he2012incremental,xu2013gosus,wang2012probabilistic},
which speedup the computation by gradually updating the low-rank structure under video background through incrementally treating only one frame at a time. Such online amelioration always significantly speeds up the calculation for the task, and makes it possible to efficiently handle the task even in real time under large-scaled video contexts.

However, the current online background subtraction techniques still have evident defects when being applied to real videos. On one hand, most current methods assume a low-rank structure for video background while neglect frequently-occurring dynamic camera jitters, such as translation, rotation, scaling and light/shade change, across video sequences. Such issues, however, always happen in real life due to camera status switching or circumstance changing over time and tend to damage the conventional low-rank assumption for video backgrounds. Actually, the image sequence formed by slightly translating/rotating/scaling each of its single images will always have no low-rank property at all. Thus the performance of current methods tend to be evidently degenerated in such background-changing cases, and it should be critical to make the online learning capable of adapting such camera jitters.

On the other hand, all current online methods for this task used a fixed loss term, e.g., $L_2$ or $L_1$ losses, in their models, which implicitly assume that noises (foregrounds) involved in videos follow a fixed probability distribution, e.g., Gaussian or Laplacian. Such assumption, however, deviates from the real scenarios where the foregrounds always have dramatic variations over time. E.g., in some frames there are no foreground objects existed, where noises can be properly modeled as a Gaussian (i.e., $L_2$-norm loss), in other cases there might be an object occluding a large area in the background, where noises should be better modeled as a long tailed Laplacian (i.e., $L_1$-norm loss), while in more often cases, the foreground might contain multiple modalities of noises, as those depicted in Fig. \ref{figure:f1}, which require to consider more complex noise models. The ignoring of such important insight of video foreground diversity always makes current methods not robust enough to finely adapt real-time foreground/noise variations in practice.

To alleviate the aforementioned issues, in this work we propose a new online background subtraction method. The contribution can be summarized as follows:

Firstly, instead of using fixed noise distribution throughout all video frames as conventional, the proposed method models the noise/foregound of each video frame as a separate mixture of Gaussian (MoG) distribution, regularized by a penalty for enforcing its parameters close to those
calculated from the previous frames. Such penalty can be equivalently reformulated as the conjugate prior, encoding the noise knowledge previous learned, for the MoG noise of current frame. Due to the good approximation capability of MoG to a wide range of distributions, our method can finely adapt video foreground variations even when the video noises are with dynamic complex structures.

Secondly, we have involved an affine transformation operator for each video frame into the proposed model, which can be automatically fitted from the temporal video contexts. Such amelioration makes our method capable of adapting wide range of video background \mbox{transformations}, like translation, rotation, scaling and any combinations of them, through properly aligning video backgrounds to make them residing on a low-rank subspace in an online manner. The proposed method can thus perform evidently more robust on the videos with dynamical camera jitters as compared with previous methods.

Thirdly, the efficiency of our model is further enhanced by embedding the sub-sampling technique into calculation. By utilizing this strategy, the proposed method can be accelerated to execute more than $250$ frames per second on average (in Matlab platform), while still keeping a good performance in accuracy, which meets the real-time requirement for practical video processing tasks. Besides, attributed to the MoG noise modeling methodology, the separated foreground layers always can be interpreted with certain physical meanings, as shown in Fig. \ref{figure:f1}, which facilitates us to get more intrinsic knowledge under video foreground.

Fourthly, our method can be easily extended to other subspace alignment tasks, like image alignment and video stabilization applications. This implies the good generalization of the proposed method.

The paper is organized as follows: Section 2 reviews some related works. Section 3 proposes our model and related algorithms. Its sub-sampling amelioration and other extensions are also introduced in this section. Section 4 shows experimental results on synthetic and real videos, to substantiate the superiority of the proposed method. Discussions and concluding remark are finally given.

\section{Related Work}\label{sec:introduction}

\subsection{Low Rank Matrix Factorization}
low rank matrix factorization (LRMF) is one of the most commonly utilized subspace learning approaches for background subtraction. The main idea is to extract the low-rank approximation of the data matrix from the product of two smaller matrices, corresponding to the basis matrix and coefficient matrix, respectively. Based on the loss terms utilized to measure the approximation extent, the LRMF methods can be mainly categorized into three classes. $L_2$-LRMF methods~\cite{de2003framework} 
utilizes $L_2$-norm loss in the model, implicitly assuming that the noise distribution in data is Gaussian. Typical $L_2$-LRMF methods include weighted SVD~\cite{gabriel1979lower}, WLRA~\cite{srebro2003weighted}, $L_2$-Wiberg~\cite{okatani2007wiberg} and so on. To make the LRMF method less sensitive to outliers, some robust loss functions have been utilized, in which the $L_1$-LRMF methods are the most typical ones. The $L_1$-LRMF utilizes the $L_1$ loss term, implying that the data noise follows a Laplacian distribution. Due to the heavy-tailed characteristic of Laplacian, such method always could perform more robust in the presence of heavy noises/outliers. Some commonly adopted $L_1$-LRMF methods include: L1Wiberg~\cite{eriksson2010efficient}, RegL1MF~\cite{zheng2012practical}, PRMF~\cite{wang2012probabilistic} 
and so on. To adapt more complex noise configurations in data, several models have recently been proposed to encode the noise as a parametric probabilistic model, and accordingly learn the loss term as well as the model parameters simultaneously. In this way, the model is capable of adapting wider range of noises as compared with the previous ones with fixed noise distributions. The typical methods in this category include the MoG-LRMF~\cite{meng2013robust,zhao2014robust} and MoEP-LRMF~\cite{cao2015moep} methods, representing noise distributions as a MoG and a mixture of exponential power distributions, respectively.
Despite having been verified to be effective in certain scenarios, these methods implicitly assume stable backgrounds across all video frames and fixed noise distribution for foreground objects throughout videos. As we have analyzed, neither is proper for practically collected videos, which tends to degenerate their performance.

\vspace{-3mm}
\subsection{Background Subtraction}
As a fundamental research topic in video processing, background subtraction has been investigated widely nowadays. The initial strategies mainly assumed that the distribution (along time) of background pixels can be distinguished from that of foreground ones. Thus by judging if a pixel is significantly deviated from the background pixel distribution, we can easily categorize if a pixel is located in background/foreground. The simplest methods along this line directly utilize a statistic measure, like the median~\cite{mcfarlane1995segmentation} or mean~\cite{lee2002background} to encode background knowledge. Later more complex distributions on background pixels, like MOG~\cite{stauffer1999adaptive}, MOGG~\cite{allili2007robust} and so on~\cite{nguyen2012robust}~\cite{haines2012background}, are more effective. The disadvantage of these methods is that they neglect useful video structure knowledge, e.g., temporal similarity of background scene and spatial contiguity of foreground objects, and thus always cannot guarantee a good performance practically. Low-rank subspace learning models represent the recent state-of-the-art for this task on general surveillance videos due to their better consideration of video structures. These methods implicitly assume stable background in videos, which are naturally with a low-rank structure. Multiple models have been raised on this topic recently, typically including PCP~\cite{candes2011robust}, GODEC~\cite{zhou2011godec}, and DECOLOR~\cite{zhou2013moving}. Albeit obtaining state-of-the-art performance in some benchmark video sets, these methods still cannot be effectively utilized in real-time problems due to both their simplified assumptions in video backgrounds (with stationary background scenes) and foreground (with fixed type of noise distributions along time). They also tend to encounter efficiency problem for real-time requirements, especially for large scaled videos. Very recently, some deep neural network works~\cite{braham2016deep,babaee2017deep} were also attempted on the task against specific scenes, while need large amount of pre-annotations. In this paper we mainly focus on handling general surveillance videos without any supervised foreground/background knowledge, and thus have not considered this approach in our experiment comparison.

\vspace{-3mm}
\subsection{Online Subspace Learning}

Nowadays, it has been attracting increasing attention to design online subspace learning method to handle real-time background subtraction issues~\cite{javed2015or,javed2015background}. The \mbox{basic idea} is to calculate only one frame \mbox{at a} time, and gradually ameliorate the background based on the real-time video variations. The state-of-the-art methods along this line include GRASTA~\cite{he2012incremental}, OPRMF~\cite{wang2012probabilistic}, GOSUS~\cite{xu2013gosus}, PracReProCS~\cite{guo2014online} and incPCP~\cite{rodriguez2016incremental,rodriguez2014matlab}. GRASTA used a $L_1$ norm loss for each frame to encode sparse foreground objects, and employed ADMM strategy for subspace updating. Similar to GRASTA, OPRMF also optimized a $L_1$-norm loss term while added regularization terms to subspace parameters to alleviate overfitting. GOSUS designed a more complex loss term to encode the structure of video foreground, and the updating algorithm is designed similar to that of GRASTA. Besides, PracReProCS and incPCP were recently proposed, which are the incremental extensions of the classical PCP algorithm.

However, these methods are still deficient due to their insufficient consideration on variations both in background and foreground in real videos. On one hand, they assume a low-rank structure for the video background, which ignores very often existed background changes and camera jitters across video sequences. On the other hand, they all fix the loss term in their models, which implicitly assumes that noise involved in data is generated from a fixed probability distribution. This, however, under-estimates the temporal variations of the foreground objects in videos. That is, in some frames the foreground signals might be very weak while in others they might be very evident. The noise distributions are thus not fixed while varying across video frames. The underestimation of both foreground/background knowledge incline to degenerate their capability for real online tasks.

\vspace{-2mm}
\subsection{Robust Subspace Alignment}

\begin{figure}[!]
\hspace{3mm}
\subfigure{
\includegraphics[width=0.45\textwidth]{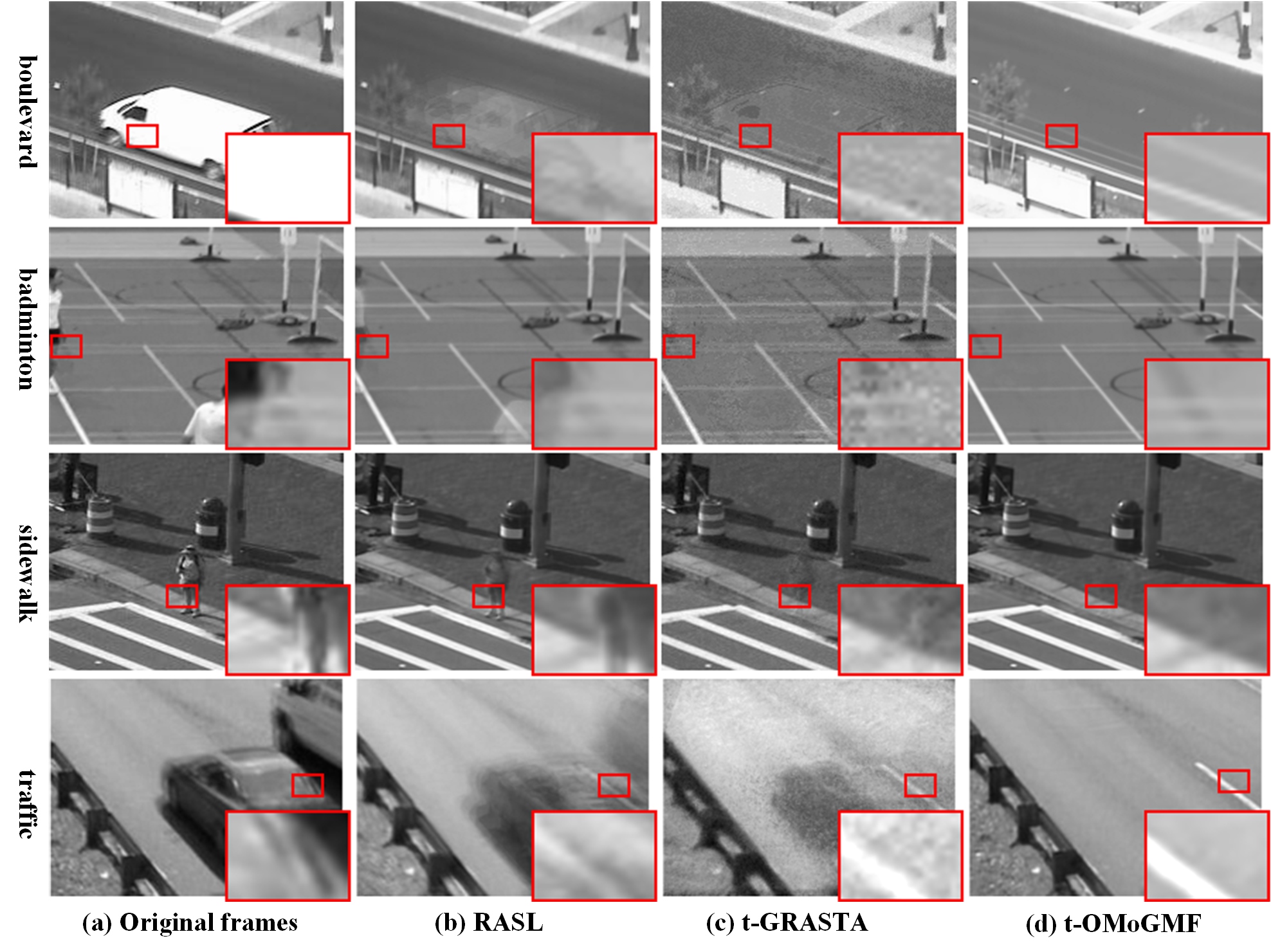}}\vspace{-5mm}
\caption{From left to right: original frames in Camera Jitter videos, backgrounds extracted by RASL, t-GRASTA and t-OMoGMF.}\vspace{-3mm}
\label{Fig2}\vspace{-0mm}
\end{figure}

Recently, multiple subspace learning strategies have been constructed to learn transformation operators on video frames to make the methods robust to camera jitters. A typical method is RASL (robust alignment by sparse and low-rank decomposition)~\cite{peng2012rasl}, which poses the learning of transformation operators into the classical robust principal component analysis (RPCA) model, and simultaneously optimize the parameters involved in such operators as well as the low-rank (background) and sparse (foreground) matrices. Other similar works are extended by
~\cite{ebadi2015approximated,ebadi2015efficient,ebadi2016dynamic}.
However, such batch-mode methods are generally slow to run and can only deal with moderate scaled videos. To this issue, incPCP\_TI~\cite{rodriguez2015translational} is extended from incPCP by taking translation and rotation into consideration to deal with image rigid transformation. t-GRASTA~\cite{he2014iterative} realized a more general subspace alignment by embedding an affine transformation operator into online subspace learning. Although capable of speeding up the offline methods, the methods utilized a simple $L_1$-norm loss to model foreground. This simple loss cannot reflect dramatic foreground variations always existed in real videos due to the fact that a simple Laplacian cannot finely reflect the complex configurations of video foregrounds. This deficiency inclines to degenerate its performance on online background subtraction. Comparatively, our proposed method fully encodes both dynamic background and foreground variations in videos, and thus is always expected to attain a better background subtraction performance, as depicted in Fig. \ref{Fig2}.

\section{Online MoG-LRMF}
We first briefly introduce the MoG-LRMF method~\cite{ meng2013robust}, which is closely related to the modeling strategy for foreground variations in our method.

\subsection{MoG-LRMF Review}
Let $\mathbf{X}=[\mathbf{x}_{1},...,\mathbf{x}_{n}]\in\Re^{d\times n}$ be the given data matrix, where $d,n$ denote the dimensionality and number of data, respectively, and each column $\mathbf{x}_{i}$ is a $d$-dimensional measurement. A general LRMF problem can be formulated as:
\vspace{-1.5mm}
\begin{equation}
\min_{\mathbf{U},\mathbf{V}}||\mathbf{W}\odot(\mathbf{X}-\mathbf{U}\mathbf{V}^T)||_{L_{p}}, \label{LRMF}
\end{equation}
where $\mathbf{U}\in\Re^{d\times r}$ and $ \mathbf{V}\in\Re^{n\times r}$ denote the basis and coefficient matrices, with $r\ll min(d,n)$, implying the low-rank property of $\mathbf{U}\mathbf{V}^T$. $\mathbf{W}$ is the indicator matrix of the same size as $\mathbf{X}$, with $w_{ij}=0$ if $x_{ij}$ is missing and $1$ otherwise. $||.||_{L_{p}}$ denotes the $p^{th}$ power of an $L_{p}$ norm, most commonly adopted as $L_{2}$ and $L_{1}$ norms in the previous research.
Eq. (\ref{LRMF}) can also be equivalently understood under the maximum likelihood estimation (MLE) framework as:
\vspace{-1.5mm}
\begin{equation}
x_{ij}=(\mathbf{u}_i)^T\mathbf{v}_j+e_{ij}, \label{eq2}
\end{equation}
where $\mathbf{u}_i,\mathbf{v}_j\in \Re^{r}$  are the $i^{th}$ and $j^{th}$ row vectors of $\mathbf{U}$ and $\mathbf{V}$, respectively, and $e_{ij}$ denotes the noise element embedded in $x_{ij}$. Under the assumption that the noise $e_{ij}$ follows a Gaussian/Laplacian distribution, the MLE model exactly complies with Eq. (\ref{LRMF}) with $L_{2}$/$L_{1}$ norm loss terms. This means the $L_2$/$L_1$-norm LRMF implicitly assume that the noise distribution underlying data follows a Gaussian/Laplacian distribution. Such simple assumption always deviates from real cases, which generally contain more complicated noise configurations~\cite{meng2013robust,cao2015moep}.

To make the model robust to complex noises, the noise term $e_{ij}$ can be modeled as a parametric probability distribution to let it more flexibly adapt real cases. Mixture of Gaussian (MoG) is naturally selected for this task~\cite{ meng2013robust} due to its strong approximation capability to general distributions. Specifically, by assuming that each  $x_{ij}$ follows
\vspace{-2.5mm}
$$
x_{ij}\sim\sum_{k=1}^K \pi_k \mathcal{N}(x_{ij}|(\mathbf{u}_i)^T\mathbf{v}_j,\sigma_k^2)
\vspace{-2.5mm}
$$
under the i.i.d. assumption, we can then get the log-likelihood function as follows:
\vspace{-0.01mm}
\begin{equation}
\begin{split}
\mathcal{L}(\mathbf{U},&\mathbf{V},\mathbf{\Pi},\mathbf{\Sigma}|\mathbf{X})=
 \sum_{i,j\in\Omega}\ln p(x_{ij}|\mathbf{\Pi},\mathbf{\Sigma},\mathbf{u}_i,\mathbf{v}_j),
\end{split}
\normalsize
\end{equation}
\vspace{-2mm}

\noindent
where $\mathbf{\Pi}=\{\pi_{k}\}_{k=1}^K$ and $\mathbf{\Sigma}=\{\sigma_{k}^2\}_{k=1}^K$ denote the mixture rates and variances involved in MoG, respectively.
The EM algorithm~\cite{dempster1977maximum} can then be readily utilized to estimate all parameters in the model, including the responsibility parameters $
\gamma_{ijk}=\frac{\pi_k\mathcal{N}(x_{ij}|(\textbf{u}_i)^T\textbf{v}_j,\sigma_k^2)}{\sum_{k=1}^K \pi_k\mathcal{N}(x_{ij}|(\textbf{u}_i)^T\textbf{v}_j,\sigma_k^2)}$ (in E-step), the MoG parameters $\pi_{k}=\frac{N_{k}}{\sum_{k=1}^KN_{k}}$, $\sigma_{k}^2=\frac{1}{N_{k}}\sum_{i,j}\gamma_{ijk}(x_{ij}-(\textbf{u}_i)^T\textbf{v}_j)^2$, where $N_{k}=\sum_{i,j}\gamma_{ijk}$, and the subspace parameters $\mathbf{U}$,$\mathbf{V}$ through solving a weighted-$L_2$-LRMF problem~\cite{meng2013robust}:
\begin{equation}
\min_{\mathbf{U},\mathbf{V}}||\mathbf{W}\odot(\mathbf{X}-\mathbf{U}\mathbf{V}^T)||_{F}, \label{WeightedL2LRMF}
\end{equation}
where $\sqrt{\sum_{k=1}^K\frac{\gamma_{ijk}}{2\sigma_k^2}}$ for $i,j\notin\Omega$, and $0$ otherwise (in M step).
Then any off-the-shelf weighted $L_2$-LRMF methods~\cite{de2003framework,srebro2003weighted} can be utilized to solve the problem.

\subsection{Online MoG-LRMF: Model}

\subsubsection{Probabilistic Modeling}

The main idea of the online MoG-LRMF (OMoGMF) method is to gradually fit a specific MoG noise distribution for foreground and a specific subspace for background for each newly coming frame $\mathbf{x}^{t}$ along the video sequence, under the regularizations of the foreground/background knowledge learned from the previous frames.
The updated MoG noise parameters include $\mathbf{\Pi}^t=\{\pi_{k}^t\}_{k=1}^K$, $\mathbf{\Sigma}^t=\{{\sigma_{k}^t}^2\}_{k=1}^K$, which are regularized under the previous learned noise knowledge $\mathbf{\Pi}^{t-1}$, $\mathbf{\Sigma}^{t-1}$ and $\{N_{k}^{t-1}\}_{k=1}^K$. The updated subspace parameters include the coefficient vector $\mathbf{v}^t$ for $\mathbf{x}^{t}$ and the current subspace $\textbf{U}^t$, required to be regularized under the previous learned subspace $\textbf{U}^{t-1}$.

\begin{figure}[!]
\centering
\subfigure{
\includegraphics[width=0.33\textwidth]{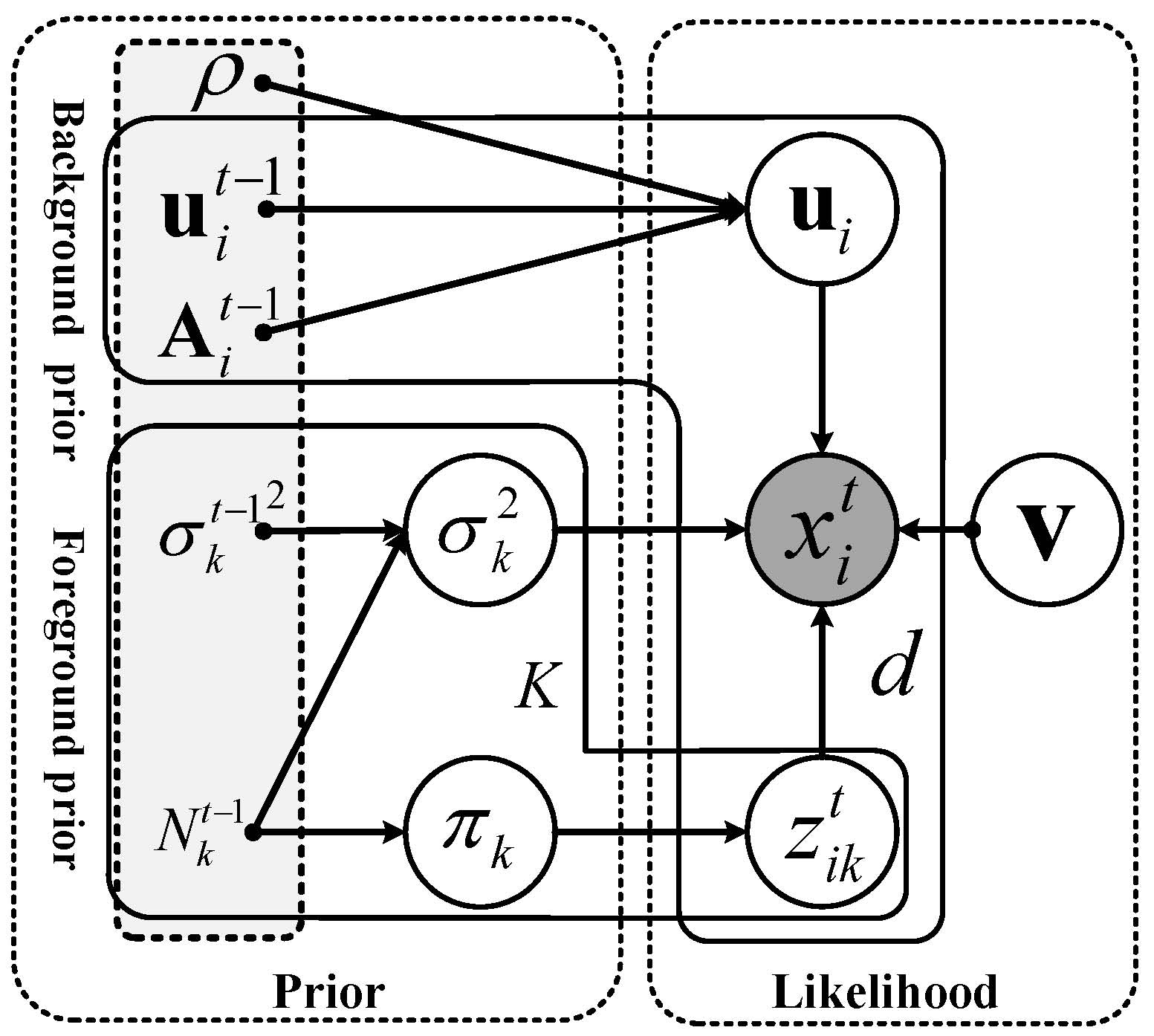}}\vspace{-3.5mm}
\caption{The graphical model for OMoGMF.}\vspace{-4mm}
\label{GraphicalModel}
\end{figure}

The model of OMoGMF can be deduced from a MAP (Maximum a posteriori) estimation by assuming a separate MoG noise distribution on each newly coming frame $\mathbf{x}^{t}$ in Eq. (\ref{eq2}), and then we have:
\vspace{-2mm}
\begin{equation}\small
\begin{split}
x_{i}^t&\sim\prod_{k=1}^K\mathcal{N}(x_{i}^t|(\mathbf{u}_i)^T\mathbf{v},\sigma_k^2)^{z_{ik}^t},\ \
\mathbf{z}_{i}^t\sim \text{Multi}(\mathbf{z}_{i}^t|\boldsymbol{\Pi}),
\end{split}\label{MoG_P}
\end{equation}\vspace{-4mm}

\noindent where $x_{i}^t$ means the $i^{th}$ pixel of $\mathbf{x}^{t}$ and Multi denotes the multinomial distribution.
We then formulate prior terms for foreground and background, respectively. For the MoG parameters of foreground,
we set the natural conjugate priors to $\sigma_k^2$ and $\boldsymbol{\Pi}$, which are the Inverse-Gamma and Dirichlet distributions~\cite{bishop2006pattern}, respectively, as follows:\vspace{-2mm}
\begin{equation}\small
\begin{split}
&\sigma_k^2\sim \text{Inv-Gamma}(\sigma_k^2|\frac{N_k^{t-1}}{2}-1,\frac{N_k^{t-1}{\sigma_k^{t-1}}^2}{2}),\\
&\boldsymbol{{\Pi}}\sim \text{Dir}(\boldsymbol{\Pi}|\boldsymbol{\alpha}),\boldsymbol{\alpha}=(N^{t-1}\pi_1^{t-1}+1,...,N^{t-1}\pi_K^{t-1}+1),
\end{split}\label{MoG_P}
\end{equation}\vspace{-1mm}

\noindent where $N^{t-1}=\sum_{k=1}^K{N_k^{t-1}}$, $\pi_k^{t-1}=N_k^{t-1}/N^{t-1}$. It can be calculated that the maximum of the above conjugate priors are $\mathbf{\Sigma}^{t-1}$ and $\mathbf{\Pi}^{t-1}$. This implies that the priors implicitly encode the previously learned noise knowledge into the OMoGMF model, and help rectify the MoG parameters of the current frame not too far from the previous learned ones.
For the subspace of background, a Gaussian distribution prior can be easily set for its each row vector:\vspace{-2mm}
\begin{equation}\small
\mathbf{u}_i\sim \mathcal{N}(\mathbf{u}_i|{\mathbf{u}_i}^{t-1},\frac{1}{\rho}\mathbf{A}_i^{t-1}),
\vspace{-2mm}
\end{equation}
where $\frac{1}{\rho}\mathbf{A}_i^{t-1}$ is a positive semi-definite matrix.
 This
facilitate the to-be-learned subspace variable $\textbf{U}$ being well regularized by the previously learned $\textbf{U}^{t-1}$. Details of how to set $\mathbf{A}_i^{t-1}$ will be introduced in Sec. 3.4.
To make a complete Bayesian model, we also set a noninformative prior $p(\mathbf{v})$ for $\mathbf{v}$, which does not intrinsically influence the calculation.
The full graphical model is depicted as Fig. \ref{GraphicalModel}.


\subsubsection{Objective Function}
All hyperparameters  are denoted by $\boldsymbol{\Theta}^{t-1}$, and after marginalizing the latent variable $\mathbf{z}^t$, we can get the posterior distribution of $\{\mathbf{\Pi},\mathbf{\Sigma},\mathbf{v},\mathbf{U}\}$
 in the following form:\vspace{-1mm}
\begin{equation}\small
\begin{split}
& p(\mathbf{\Pi},\mathbf{\Sigma},\mathbf{v},\mathbf{U}|\mathbf{x}^t,\boldsymbol{\Theta}^{t-1})\propto\\
&\ \ \ \ \ p(\mathbf{x}^t|\mathbf{\Pi},\mathbf{\Sigma},\mathbf{v},\mathbf{U})p(\mathbf{\Sigma}|\boldsymbol{\Theta}^{t-1})
p(\mathbf{\Pi}|\boldsymbol{\Theta}^{t-1})p(\mathbf{U}|\boldsymbol{\Theta}^{t-1})p(\mathbf{v}).
\end{split}\label{MAP}
\end{equation}
Based on the MAP principle, we can get the following minimization problem for calculating $\mathbf{\Pi}^t, \mathbf{\Sigma}^t, \mathbf{v}^t,\mathbf{U}^t$:
\begin{equation}
\footnotesize
\begin{split}
\mathcal{L}^t(\mathbf{\Pi},\mathbf{\Sigma},\mathbf{v},\mathbf{U})=
-\ln p(\mathbf{x}^t|\mathbf{\Pi},\mathbf{\Sigma},\mathbf{v},\mathbf{U})
+\mathcal{R}_F^t(\mathbf{\Pi},\mathbf{\Sigma})+
\mathcal{R}_B^t(\mathbf{U}),
\end{split}\label{MoGLRMF1}
\end{equation}
where\vspace{-3mm}
\begin{equation*}
\small
\begin{split}
&\ln p(\mathbf{x}^t|\mathbf{\Pi},\mathbf{\Sigma},\mathbf{v},\mathbf{U})=
\sum_{i}\ln(\sum_{k=1}^K \pi_k\mathcal{N}(x_{i}^t|(\mathbf{u}_i)^T\mathbf{v},\sigma_k^2)),\\
&\mathcal{R}_F^t(\mathbf{\Pi},\mathbf{\Sigma})=\sum_{k=1}^KN_{k}^{t-1}(
\frac{1}{2}\frac{{\sigma_k^{t-1}}^2}{\sigma_k^2}+\ln\sigma_k)
-N^{t-1}\sum_{k=1}^K{\pi_{k}^{t-1}\ln\pi_{k}},\\
&\mathcal{R}_B^t(\mathbf{U})=\rho\sum_{i=1}^{d}({\mathbf{u}_i}-{\mathbf{u}_i}^{t-1})^T
{(\mathbf{A}_i^{t-1}})^{-1}({\mathbf{u}_i}-{\mathbf{u}_i}^{t-1}).
\end{split}\label{MoGLRMF1_1}
\vspace{-1mm}
\end{equation*}\vspace{-3mm}

\noindent In the above problem, the first term is the likelihood term, which enforces the learned parameters adapt to the current frame $\mathbf{x}^t$. The second term $\mathcal{R}_F^t(\mathbf{\Pi},\mathbf{\Sigma})$ is the regularization term
for noise distribution, whose function can be more intuitively interpreted by the following equivalent form:\vspace{-2mm}
\begin{equation}
\small
\begin{split}
{\mathcal{R}}_F^t(&\mathbf{\Pi},\mathbf{\Sigma})=
 \sum_{k=1}^K{N_{k}^{t-1}
D_{KL}(\mathcal{N}(x|0,{\sigma_k^{t-1}}^2)||\mathcal{N}(x|0,{\sigma_k}^2))}\\&\ \ \ \ \  \ \ \ +N^{t-1}D_{KL}(
\text{Multi}(\mathbf{z}|\boldsymbol{\Pi}^{t-1})||\text{Multi}(\mathbf{z}|\boldsymbol{\Pi}))+C\\
&=N^{t-1}D_{KL}(p(x,\mathbf{z}|\mathbf{\Pi}^{t-1},\mathbf{\Sigma}^{t-1})||p(x,\mathbf{z}|\mathbf{\Pi},\mathbf{\Sigma}))+C,
\end{split}\label{MoGLRMF2}
\end{equation}\vspace{-2mm}

\noindent where $\footnotesize{p(x,\mathbf{z}|\mathbf{\Pi},\mathbf{\Sigma})=\prod_{k=1}^K{{\pi_k}^{z_k}\mathcal{N}(x|0,{\sigma_k}^2)^{z_k}}}$, $\footnotesize{D_{KL}(\cdot||\cdot)}$ denotes the KL divergence between two distributions.
It can be evidently observed that $\mathcal{R}_F^t(\mathbf{\Pi},\mathbf{\Sigma})$ functions to rectify the foreground distribution on the current $t^{th}$ frame (with parameters $\mathbf{\Pi},\mathbf{\Sigma}$) to approximate the previously learned one (with parameters $\mathbf{\Pi}^{t-1},\mathbf{\Sigma}^{t-1}$).
Besides, the third term $\mathcal{R}_B^t(\mathbf{U})$ in (\ref{MoGLRMF1}) corresponds to a Mahalanobis distance between each row vector of $\mathbf{U}$ to that of $\mathbf{U}^{t-1}$, thus functioning to rectify the current learned subspace by the previously learned one. The compromising parameter $N^{t-1}$ and $\rho$ control the strength of the priors, and their physical meanings and setting manners will be introduced in Sec. 3.4.

To easily compare differences of our model with the previous ones, we list typical models along this research line, as well as ours, in Table \ref{table:model}.

\begin{table*}
\scriptsize
\caption{Model comparison of typical subspace-based background subtraction methods }
\newcommand{\tabincell}[2]{\begin{tabular}{@{}#1@{}}#2\end{tabular}}  %
\vspace{-3mm}\hspace{-5mm}
\begin{tabular}{c c c c c c}
\shline
Method &\tabincell{c}{Foreground/Background\\Decomposition }
&Objective Function&Constraint/Basic Assumption&Implementation Scheme\\
\shline
RPCA ~\cite{candes2011robust}&$\mathbf{X}=\mathbf{L}+\mathbf{S}$&
$\min_{\mathbf{L},\mathbf{S}}||\mathbf{L}||_*+\lambda||\mathbf{S}||_1$&No&Offline\\
GODEC ~\cite{zhou2011godec}&$\mathbf{X}=\mathbf{L}+\mathbf{S}+\mathbf{E}$&
$\min_{\mathbf{L},\mathbf{S}}||\mathbf{E}||_F^2$&$rank(\mathbf{L})\leq K, card(\mathbf{S})\leq s$ &Offline\\
RegL1~\cite{zheng2012practical} &$\mathbf{X}=\mathbf{U}\mathbf{V}^T+\mathbf{S}$
&$\min_{\mathbf{U},\mathbf{V}}||\mathbf{S}||_1+\lambda||\mathbf{V}||_*$&$\mathbf{U}^T\mathbf{U}=\mathbf{I}$   &Offline\\
PRMF~\cite{wang2012probabilistic}&$\mathbf{X}=\mathbf{U}\mathbf{V}^T+\mathbf{E}$&
$\min_{\mathbf{U},\mathbf{V}}-\ln p(\mathbf{X}|\mathbf{U},\mathbf{V})+\lambda_1||\mathbf{U}||_F^2+\lambda_2||\mathbf{V}||_F^2$&
$e_{ij}\sim \mathcal{L}(e|0,\lambda)$
&Offline\\
DECOLOR~\cite{zhou2013moving}&$\mathbf{X}=\mathbf{L}+\mathbf{E}$
&\tabincell{c}{$\min_{\mathbf{L},\mathbf{S}}||\mathbf{S}^\perp\odot\mathbf{E}||_F^2+\lambda_1||\mathbf{L}||_*$\\$+\lambda_2||\mathbf{S}||_1
+\lambda_3||\mathbf{S}||_{TV}$}
&$s_{ij}\in\{0,1\}$&Offline\\
\hline
GRASTA~\cite{he2012incremental}&$\mathbf{x}^t=\mathbf{U}\mathbf{v}+\mathbf{s}$&
$\min_{\mathbf{v}}||\mathbf{s}||_1$&$\mathbf{U}^T\mathbf{U}=\mathbf{I}$&Online: Heuristically update $\mathbf{U}$\\
OPRMF~\cite{wang2012probabilistic}&$\mathbf{x}^t=\mathbf{U}\mathbf{v}+\mathbf{e}$&
$\min_{\mathbf{v}}-\ln p(\mathbf{x}^t|\mathbf{U},\mathbf{v})+\lambda||\mathbf{v}||_2^2$
&$e_{i}\sim \mathcal{L}(e|0,\lambda)$&Online: Heuristically update $\mathbf{U}$\\
GOSUS~\cite{xu2013gosus}&$\mathbf{x}^t=\mathbf{U}\mathbf{v}+\mathbf{s}+\mathbf{e}$&
$\min_{\mathbf{v}}{||\mathbf{e}||_2^2+\sum_{l=1}^L\lambda_l{||\mathbf{D}^l\mathbf{s}||_2}}$
&$\mathbf{U}^T\mathbf{U}=\mathbf{I}$&Online: Heuristically update $\mathbf{U}$\\
\textbf{OMoGMF}&$\mathbf{x}^t=\mathbf{U}\mathbf{v}+\mathbf{e}$&
\tabincell{c}{$\min_{\mathbf{\Pi},\mathbf{\Sigma},\mathbf{v},\mathbf{U}}
-\ln p(\mathbf{x}^t|\mathbf{\Pi},\mathbf{\Sigma},\mathbf{v},\mathbf{U})$\\$+
\mathcal{R}_F^t(\mathbf{\Pi},\mathbf{\Sigma})+\mathcal{R}_B^t(\mathbf{U})$\\ }
&$e_{i}\sim \sum_{k=1}^K \pi_k\mathcal{N}(e|0,\sigma_k^2)$&Online: Optimize $\mathbf{U}$\\
\shline
\end{tabular}\vspace{-3mm}
\label{table:model}
\end{table*}

\subsection{Online MoG-LRMF: Algorithm}

The online-EM algorithm can be readily utilized for solving the OMoGMF model (\ref{MoGLRMF1}), by alternatively implementing
the following E-step and M-step on a new frame sample $\mathbf{x}^{t}$.

\textbf{Online E Step}: As the traditional EM strategy, this step aims to estimate the expectation of posterior probability for
latent variable $z_{ik}^{t}$, which is also known as
 responsibility $\gamma_{ik}^{t}$.
 The updating equation is as follows:
\begin{equation}
E(z_{ik}^{t})=\gamma_{ik}^{t}=\frac{\pi_k\mathcal{N}(x_{i}^{t}|(\textbf{u}_i)^T\textbf{v},{\sigma_k}^2)}{\sum_{k=1}^K \pi_k\mathcal{N}(x_{i}^{t}|(\textbf{u}_i)^T\textbf{v},{\sigma_k}^2)}.
\label{OES}
\end{equation}

\textbf{Online M Step}: On updating MoG parameters $\mathbf{\Pi},\mathbf{\Sigma}$,
we need to minimize the following sub-optimization problem:
\begin{equation}
\small
\begin{split}
\mathcal{L'}^t(\mathbf{\Pi},\mathbf{\Sigma})=
-E_{\mathbf{z}^t}\{\ln p(\mathbf{x}^t,\mathbf{z}^t|\mathbf{\Pi},\mathbf{\Sigma},\mathbf{v},\mathbf{U})\}
+\mathcal{R}_F^t(\mathbf{\Pi},\mathbf{\Sigma})
\end{split}.\label{MoGLRMF_LB}
\end{equation}
The closed-form solution is\footnote{Inference details are listed in the supplementary material (SM).}:
\begin{equation}\small
\begin{split}
&\pi_k=\pi_k^{t-1}-\frac{\overline{N}}{N}(\pi_k^{t-1}-\overline{\pi}_k);\\
&{\sigma_{k}}^2={\sigma_{k}^{t-1}}^2-\frac{\overline{N}_{k}}{N_{k}}({\sigma_{k}^{t-1}}^2-{\overline{\sigma}_{k}}^2)
\end{split}\label{OMS1}
\end{equation}
where \vspace{-5mm}
\begin{equation}\small
\begin{split}
&\overline{N}=d;\  \overline{N}_{k}=\sum_i^d\gamma_{ik}^t; \ \overline{\pi}_k=\frac{\overline{N}_{k}}{\overline{N}};\\
&{\overline{\sigma}_{k}}^2=\frac{1}{\overline{N}_{k}}\sum_{i=1}^d\gamma_{ik}^t(x_{i}^{t}-(\textbf{u}_i)^T\textbf{v})^2);\\
&N=N^{t-1}+\overline{N}; N_k=N^{t-1}_k+\overline{N}_k.
\end{split}\label{OMS2}
\end{equation}

On updating coefficient parameter $\textbf{v}$, we need to solve the sub-optimization problem of (\ref{MoGLRMF1}) with respect to $\textbf{v}$ as:
\begin{equation}
\begin{aligned}
&\min_{\mathbf{v}}||\mathbf{w}^t\odot(\mathbf{x}^{t}-\mathbf{U}\mathbf{v})||_{F}^2,
\end{aligned}\label{UpdateV}
\end{equation}\vspace{-4mm}

\noindent
where each element of $\mathbf{w}^t$ is $w_i^t=\sqrt{\sum_{k=1}^K\frac{\gamma_{ik}^{t}}{2{\sigma_k}^2}}$ for $i=1,2,...,d$. This problem is a weighted least square problem, and has the closed-form solution as:
\begin{equation}
\mathbf{v}=({\mathbf{U}}^Tdiag(\mathbf{w}^t)^2
\mathbf{U})^{-1}{\mathbf{U}}^Tdiag(\mathbf{w}^t)^2\mathbf{x}^{t}. \label{updateV-2}
\end{equation}

On updating the subspace parameter $\mathbf{U}$, we need to solve the following sub-problem of (\ref{MoGLRMF1}):
\vspace{-1mm}
\begin{equation}\small
\begin{split}
\small
&\mathcal{L'}^t(\mathbf{U})=
-E_{\mathbf{z}^t}\{\ln p(\mathbf{x}^t,\mathbf{z}^t|\mathbf{\Pi},\mathbf{\Sigma},\mathbf{v},\mathbf{U})\}
+\mathcal{R}_B^t(\mathbf{U})\\
&=||\mathbf{w}^{t}\odot(\mathbf{x}^{t}-\mathbf{U}{\mathbf{v}^{t}})||_{F}^2+\mathcal{R}_b^t(\mathbf{U}),
\label{NewWeightedL2LRMF}
\end{split}
\end{equation}
and it has closed-form solution for each its row vector as:\vspace{-1mm}
\begin{equation*}
\footnotesize
\mathbf{u}_i^{t}={\left (\rho{(\mathbf{A}_i^{t-1}})^{-1}+{w_{i}^{t}}^2
\mathbf{v}^{t}{\mathbf{v}^{t}}^T\right )}^{-1}(\rho{(\mathbf{A}_i^{t-1}})^{-1}
\mathbf{u}_i^{t-1}+{w_{i}^{t}}^2x_i^t{\mathbf{v}^{t}}^T).
\vspace{-1mm}
\end{equation*}

In order to get a simple updating rule, we set
\vspace{-1mm}
\begin{equation}\small
\begin{split}
&({\mathbf{A}_i^{t}})^{-1}=\rho({\mathbf{A}_i^{t-1}})^{-1}+{w_{i}^{t}}^2
\mathbf{v}^{t}{\mathbf{v}^{t}}^T; \\ &{\mathbf{b}_i^{t}}=\rho({\mathbf{A}_i^{t-1}})^{-1}
{\mathbf{u}_i}^{t-1}+{w_{i}^{t}}^2x_i^t{\mathbf{v}^{t}}^T,
\end{split}\label{updateAb1}
\end{equation}
\noindent and then we have ${\mathbf{u}_i}^{t}=\mathbf{A}_i^{t}\mathbf{b}_i^{t}$. By using matrix inverse equation~\cite{bishop2006pattern} and the equation
 ${\mathbf{u}_i}^{t-1}=\mathbf{A}_i^{t-1}\mathbf{b}_i^{t-1}$, the update rules for $\mathbf{A}_i^{t}$ and $\mathbf{b}_i^{t}$ can be reformulated as:
\begin{equation}\small
\begin{aligned}
\mathbf{A}_i^{t}&=\frac{1}{\rho}\left (\mathbf{A}_i^{t-1}-\frac{{w_{i}^{t}}^2
\mathbf{A}_i^{t-1}\mathbf{v}^{t}{\mathbf{v}^{t}}^T\mathbf{A}_i^{t-1}}
{\rho+{w_{i}^{t}}^2{\mathbf{v}^{t}}^T\mathbf{A}_i^{t-1}\mathbf{v}^{t}}\right );\\
\mathbf{b}_i^{t}&=\rho\mathbf{b}_i^{t-1}+{w_{i}^{t}}^2x_i^{t}\mathbf{v}^{t}.
\end{aligned}\label{updateAb}
\end{equation}
Thus in each step of updating $\mathbf{U}^t$, we only need to save $\{\mathbf{A}_i^{t-1}\}_{i=1}^d$,  $\{\mathbf{b}_i^{t-1}\}_{i=1}^d$ calculated in the last step, which only needs fixed storage memory. Note that since the matrix inverse computations are avoided in the above updating equations, the efficiency of the algorithm is guaranteed.

Since the subspace, representing the background knowledge, changes relatively slowly along the video sequence, we only fine-tune $\mathbf{U}$ once after recursively implementing the above E-M steps on updating $\{\gamma_{ik}^{t}\}_{ik}$, $\mathbf{\Pi}^{t}$, $\mathbf{\Sigma}^{t}$, and $\textbf{v}^{t}$  until convergence for each new sample $\mathbf{x}^t$ under fixed subspace $\mathbf{U}^{t-1}$. The subspace can then be fine-tuned to adjust the temporal background change in this video frame. Note that there are only simple computations involved in the above updating process, except that in (\ref{updateV-2}), we need to compute the inverse of a $r\times r$ matrix. In the background subtraction contexts, the rank $r$ is generally with a small value and far less than $d,n$. We thus can very efficiently calculate this matrix inverse in general.

The OMoGMF algorithm can then be summarized in $\mathbf{Algorithm}$ $\mathbf{1}$. About initialization, we need a warm-start for starting our algorithm by running PCA on a small batch of starting video frames to get an initial subspace, employing MoG algorithm on the extracted noise to get initial MoG parameters, and calculating the initial $\{\mathbf{A}_i\}_{i=1}^d$, $\{\mathbf{b}_i\}_{i=1}^d$ for subspace learning.

\begin{algorithm}[!tbp]\small
\caption{[OMoGMF] online MoG-LRMF}\label{alg1}
\begin{algorithmic}[1]
\renewcommand{\algorithmicrequire}{\textbf{Input:}}
\renewcommand{\algorithmicensure}{\textbf{End}}
\REQUIRE the MoG parameters: $\{\mathbf{\Pi}^{t-1}, \mathbf{\Sigma}^{t-1}, N^{t-1}\}$; model variables: $\{\mathbf{A}_i^{t-1}\}_{i=1}^d$, $\{\mathbf{b}_i^{t-1}\}_{i=1}^d$, $\mathbf{U}^{t-1}$; data: $\mathbf{x}^{t}$
\renewcommand{\algorithmicrequire}{\textbf{Initialization:}}
\renewcommand{\algorithmicensure}{\textbf{End}}
\REQUIRE $\{\mathbf{\Pi}, \mathbf{\Sigma}\}=\{\mathbf{\Pi}^{t-1}, \mathbf{\Sigma}^{t-1}\}$, $\mathbf{v}^{t}$
\WHILE {not converged}
\STATE
\textbf{Online E-step}: compute  $\gamma_{ik}^t$ by (\ref{OES})
\STATE
 \textbf{Online M-step}: compute $\{\mathbf{\Pi}, \mathbf{\Sigma}, N\}$ by (\ref{OMS1}) and $\mathbf{v}$ by (\ref{updateV-2})
\ENDWHILE
\FOR{each $\mathbf{u}_i^{t}$, $i=1,2,...,d$}
\STATE
compute $\{\mathbf{A}_i^{t}\}_{i=1}^d$, $\{\mathbf{b}_i^{t}\}_{i=1}^d$ by (\ref{updateAb})
\STATE
compute $\mathbf{u}_i^{t}$ by $\mathbf{u}_i^{t}=\mathbf{A}_i^{t} \mathbf{b}_i^{t}$
\ENDFOR
\renewcommand{\algorithmicrequire}{\textbf{Output:}}
\renewcommand{\algorithmicensure}{\textbf{End}}
\REQUIRE $\{\mathbf{\Pi}^{t}, \mathbf{\Sigma}^{t}, N^{t}\}$, $\mathbf{v}^{t}$,
 $\{\mathbf{A}_i^{t}\}_{i=1}^d$, $\{\mathbf{b}_i^{t}\}_{i=1}^d$, $\mathbf{U}^{t}$.
\end{algorithmic}
\end{algorithm}


\subsection{Several remarks}
\textbf{On relationship between conjugate prior and KL divergence}:
Actually we can prove a general result to understand the conjugate prior as an equivalent KL divergence regularization. For the fully exponential family distributions, we have the following theorem\footnote{All proofs are presented in SM due to page limitation.}:\\
\textbf{Theorem 1}
\emph{If a distribution $p(\mathbf{x}|\boldsymbol{\theta})$ belongs to the full exponential family with the form:
$p(\mathbf{x}|\boldsymbol{\theta})=\eta(\boldsymbol{\theta})\text{exp}
({\boldsymbol{\theta}}^T\boldsymbol{\phi}(\mathbf{x})),$
and its conjugate prior follows: $\small{p(\boldsymbol{\theta}|\boldsymbol{\mathcal{X}},\gamma)
=f(\boldsymbol{\mathcal{X}},\gamma)\eta(\boldsymbol{\theta})^\gamma\text{exp}(\gamma
{\boldsymbol{\theta}}^T\boldsymbol{\mathcal{X}})},$
then we have:
$$\ln p(\boldsymbol{\theta}|\boldsymbol{\mathcal{X}},\gamma)=
-\gamma D_{KL}(p(\mathbf{x}|\boldsymbol{\theta}^{*})||p(\mathbf{x}|\boldsymbol{\theta}))+C,$$
where $\boldsymbol{\theta}^{*}=arg\max_{\boldsymbol{\theta}}{p(\boldsymbol{\theta}|\boldsymbol{\mathcal{X}},\gamma)}$
and $C$ is a constant independent of $\boldsymbol{\theta}$}.

Since both Gaussian and multinomial distributions belong to the full exponential family, both conjugate priors in (\ref{MoG_P}) can be written in their equivalent KL divergence expressions (\ref{MoGLRMF2}). We prefer to use the latter form in our study since it can more intuitively deliver the noise regularization insight underlying our model in a deterministic manner.

\textbf{On relationship to batch-mode model}:
Under the model setting of (\ref{MoGLRMF1}) (especially for the two regularization terms $\mathcal{R}_F^t(\mathbf{\Pi},\mathbf{\Sigma})$ and $\mathcal{R}_B^t(\mathbf{U})$), there is an intrinsic relationship between our online model incrementally implemented on current sample $\mathbf{x}^t$ with a batch-mode one on all learned samples $\{\mathbf{x}^j\}_{j=1}^t$, as described in the following theorem:
\\
\textbf{Theorem 2}
\emph{ By setting $N^{t-1}=(t-1)d$ and $\rho=1$, minimizing
(\ref{MoGLRMF_LB}) for $\{\mathbf{\Pi}, \mathbf{\Sigma}\}$ and (\ref{NewWeightedL2LRMF}) for $\mathbf{U}$
are equivalent to calculating:
\begin{equation}\small
\begin{split}
\{\mathbf{\Pi}^t, \mathbf{\Sigma}^t\}&=arg\max_{\mathbf{\Pi}, \mathbf{\Sigma}}\sum_{j=1}^t \ln p(\mathbf{x}^j,\mathbf{z}^j|\mathbf{\Pi},\mathbf{\Sigma},\mathbf{v}^j,\mathbf{U}^j),\\
\mathbf{U}^t&=arg\max_{\mathbf{U}}\sum_{j=1}^t \ln p(\mathbf{x}^j,\mathbf{z}^j|\mathbf{\Pi}^j,\mathbf{\Sigma}^j,\mathbf{v}^j,\mathbf{U}),
\end{split}\label{BatchUN}
\end{equation}
respectively.
Moreover, under these settings, it holds that:\vspace{-1mm}
\begin{equation}
\begin{split}
&||\mathbf{\Sigma}^t-\mathbf{\Sigma}^{t-1}||_F\leq O(\frac{1}{t}),
\  \ \ ||\mathbf{\Pi}^t-\mathbf{\Pi}^{t-1}||_F\leq O(\frac{1}{t}),\\
&||\mathbf{U}^t-\mathbf{U}^{t-1}||_F\leq O(\frac{1}{t}).
\end{split}
\end{equation}
}
The above result demonstrates the batch-mode understanding of our online learning schemes, under fixed previously learned variables
($\{\mathbf{z}^j,\mathbf{v}^j,\mathbf{U}^j,\mathbf{\Pi}^j,\mathbf{\Sigma}^j\}_{j=1}^{t-1}$),
which have not been stored in memory in the online implementation manner and cannot be re-calculated on previous frames.

\begin{figure}[!]
\subfigure{
\hspace{4mm}
\includegraphics[width=0.40\textwidth]{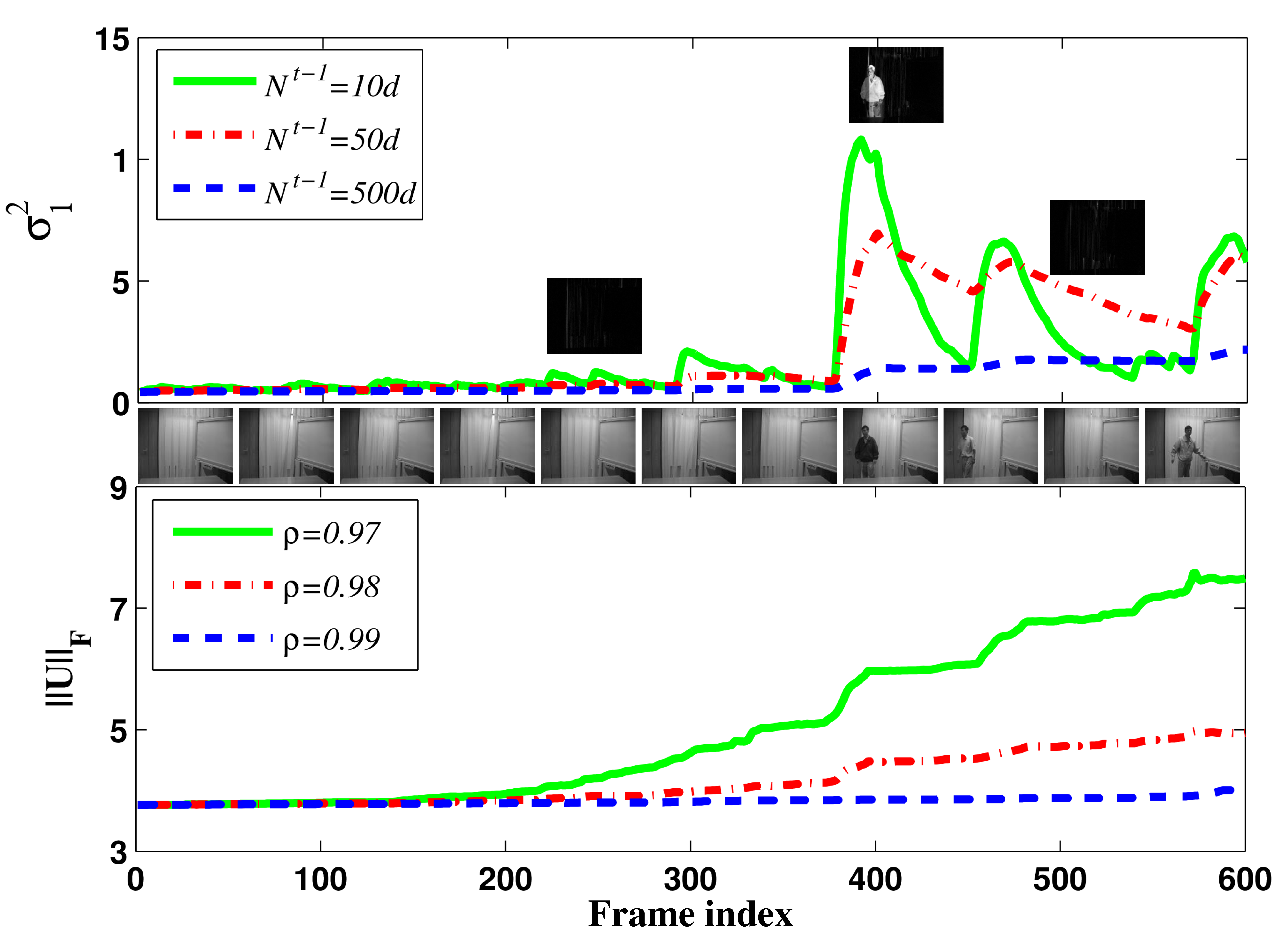}}\vspace{-4mm}
\caption{Tendency curves of the largest variance $({\sigma_{1}^{t}})^2$(with scale $10^{-2}$) and  $||\mathbf{U}^{t}||_F$ (with scale $10^{3}$) along time under different values of $N^{t-1}$ and $\rho$, respectively, for
$curtain$ sequence. Typical video frames and some foregrounds extracted by our method along time are also depicted.
}
\label{figure:MoGPara}\vspace{-2mm}
\end{figure}

{\textbf{On parameters $N^{t-1}$ and $\rho$ :}} Although
the natural choices for them are $N^{t-1}=(t-1)d$ and $\rho=1$ based on Theorem 2, under these settings, the prior knowledge learned from previous frames will be gradually accumulated (note that the value of $N^{t-1}$ will increase to infinity), and the function of the likelihood term (i.e., the effect of the current frame) will be more and more alleviated with time. However, as the motivation of this work, we expect that our method can consistently fit the foreground variations and dynamically adapt the noise changes with time, and thus hope that the likelihood term can constantly play roles in the computation. In our algorithm, we just easily set $N^{t-1}$ as a fixed constant $Kd$ ($N_k^{t-1}=N^{t-1}\pi_{k}^{t-1}$ correspondingly), meaning that we dominate the adjacent $K$ frames to rectify the online parameter updating of the current frame. In practical cases, a moderate $K$ (e.g., we set it as $50$ in all our experiments) is preferred to make the method adaptively reflect temporal variations of video foreground, while not too sensitive to single frame change, as clearly depicted in Fig. \ref{figure:MoGPara}. Similarly, we easily set $\rho$ as $0.98$ throughout our experiments to let the updated subspace slightly lean to the current frame.


\subsection{Efficiency and Accuracy Amelioration}
We then introduce two useful techniques to further enhance the efficiency and accuracy of the proposed method.

\subsubsection{Sub-sampling}
It can be shown that a large low-rank matrix can be reconstructed from a small number of its entries~\cite{candes2009exact} under certain low-rank assumption. Inspired by some previous attempts~\cite{he2012incremental} on this issue, we also prefer to use sub-sampling technique to further improve efficiency of our method.

For a newly coming frame $\mathbf{x}$, we randomly sample some of its entries to get the sub-sampling data $\mathbf{x}_\Omega$, where $\Omega$ is the index set of the sampled entries, and then we only use $\mathbf{x}_\Omega$ to update the parameters involved in our model. The updating of MoG parameters $\mathbf{\Pi}$ and $\mathbf{\Sigma}$ is similar to the original method, and $\mathbf{v}$ can be solved under the sampled subspace $\mathbf{U_\Omega}$. While for $\mathbf{U}$, we only need to update its row entries on $\Omega$ through using $\{\mathbf{A}_{i}\}_{i\in\Omega}$ and $\{\mathbf{b}_{i}\}_{i\in\Omega}$.

Generally speaking, the sub-sampling rate is inversely proportional to the performance of our method, and we thus need to find a trade-off between efficiency and accuracy in real cases. E.g., in evidently low-rank background cases, the sampling rate should be larger while for scenes with complex backgrounds across videos, we need to sample more data entries to guarantee accuracy.

\subsubsection{TV-norm regularization}
The foreground is defined as any objects which are occluded before the background during a period of time. In real-world scenes, as we know, one foreground object often appears in a continuous shape and the region of one object generally is with an evident spatial smoothness. In our online method, we also consider such spatial smoothness to further improve its accuracy on foreground object detection.

There are several strategies to encode the smoothness property of an image, e.g., Markov random field (MRF)~\cite{wang2013bayesian,senior2011interactive,su2011over}, Total Variation (TV) minimization~\cite{cao2015novel,cao2015total}, and structure sparsity~\cite{xu2013gosus,huang2011learning}. Considering effectiveness and efficiency, we employ the TV-minimization approach in our method. For a foreground frame obtained by
our method, we calculate the following TV-minimization problem:
\vspace{-1.5mm}
\begin{equation}
\vspace{-1.5mm}
\mathbf{FG}'=arg\min_\mathbf{F} \frac{1}{2}||\mathbf{F}-\mathbf{FG}||_{2}^2+\lambda||\mathbf{F}||_{TV},\label{FG}
\end{equation}
where $||\cdot||_{TV}$ is the TV norm and $\mathbf{FG}$ denotes the foreground got by our method. The optimization problem (\ref{FG}) can be readily solved by TV-threshold algorithm~\cite{wang2014highly,yang2013efficient}.
 In our experiments, we just empirically set $\lambda$ as about $1.5\widetilde{\sigma}^2$ ($\widetilde{\sigma}^2$ is the largest variance among MoG components), and our method can perform well throughout all our experiments.

\subsection{Transformed Online MoG-LRMF}
\begin{figure*}[!]
\centering
\subfigure{
\includegraphics[width=0.7\textwidth]{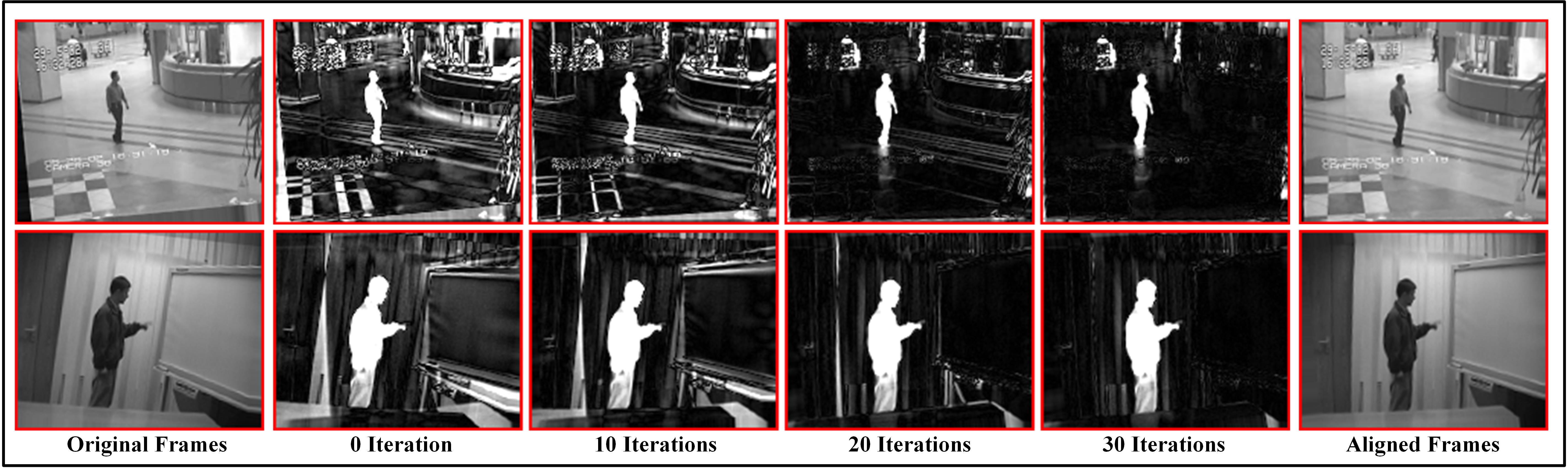}}\vspace{-3mm}
\caption{Residuals in different iterations of t-OMoGMF on a transformed frame of ``\emph{airport}" (first row)/``\emph{curtain}" (second row) video in Li dataset.}
\label{Fig3}
\end{figure*}

Due to camera jitter or circumstance changes, videos collected from real scenes always contain background changes over time, which tends to hamper the low-rank assumption on subspace learning models. A real-time alignment is thus required to guarantee the soundness of the online subspace learning methods on the background subtraction task.
To this aim, we embed a transformation operator into our model and optimize its parameters as well as other subspace and noise parameters to facilitate our model adaptable to such misaligned videos. Specifically, for a newly coming frame $\mathbf{x}$, we aim to learn a transformation operator $\boldsymbol{\tau}$ under the current subspace $\mathbf{U}$. Denote the MoG noise as
\vspace{-2mm}
\begin{equation}
\vspace{-2mm}
p(e_i)=\sum_{k=1}^K \pi_k\mathcal{N}(e_i|0,{\sigma_k}^2),i=1,...,d,
\end{equation}
where $\boldsymbol{e}=\mathbf{x}\circ\boldsymbol{\tau}-\mathbf{U}\mathbf{v}$, $e_i$ is the $i$th entry of the vector $\boldsymbol{e}$ and $\mathbf{x}\circ\boldsymbol{\tau}$ denotes the transformation with parameters $\boldsymbol{\tau}$ on $\mathbf{x}$. The transformation can be an affine or projective transformation.
\begin{algorithm}[!tbp]\small
\caption{[t-OMoGMF] transformed Online MoGMF }\label{alg2}
\begin{algorithmic}[1]
\renewcommand{\algorithmicrequire}{\textbf{Input:}}
\renewcommand{\algorithmicensure}{\textbf{End}}
\REQUIRE the MoG parameters: $\{\mathbf{\Pi}^{t-1}, \mathbf{\Sigma}^{t-1}, N^{t-1}\}$, model variables: $\{\mathbf{A}_i^{t-1}\}_{i=1}^d$, $\{\mathbf{b}_i^{t-1}\}_{i=1}^d$, $\mathbf{U}^{t-1}$, data: $\mathbf{x}^{t}$
\renewcommand{\algorithmicrequire}{\textbf{Initialization:}}
\renewcommand{\algorithmicensure}{\textbf{End}}
\REQUIRE $\{\mathbf{\Pi}, \mathbf{\Sigma}\}=\{\mathbf{\Pi}^{t-1}, \mathbf{\Sigma}^{t-1}\}$ , $\mathbf{v}$ ,$\boldsymbol{\tau}$
\WHILE {not converged}
\STATE
 Estimate the Jacobian matrix:
   $\mathbf{J}=\frac{\partial(\mathbf{x}^{t}\circ\boldsymbol{\zeta})}{\partial\boldsymbol{\zeta}}|_{\boldsymbol{\zeta}=\boldsymbol{\tau}}$
\WHILE {not converged}
\STATE
\textbf{Online E-step}: compute  $\gamma_{ik}^t$ by (\ref{OES})
\STATE
\textbf{Online M-step}: compute the MoG parameters\\ $\{\mathbf{\Pi},\mathbf{\Sigma}, N\}$ by Eq. (\ref{OMS1}) and
compute $\{\mathbf{v},\Delta\boldsymbol{\tau}\}$ by (\ref{UpdateTau})
\ENDWHILE
\STATE
Update the transformation parameters:\\
 $\boldsymbol{\tau}=\boldsymbol{\tau}+\Delta\boldsymbol{\tau}$
\ENDWHILE
\FOR{each $\mathbf{u}_i^{t}$, $i=1,2,...,d$}
\STATE
Update $\{\mathbf{A}_i^{t}\}_{i=1}^d$, $\{\mathbf{b}_i^{t}\}_{i=1}^d$ by subspace update rule (\ref{updateAb})
\STATE
Update $\mathbf{u}_i^{t}$ by $\mathbf{u}_i^{t}=\mathbf{A}_i^{t} \mathbf{b}_i^{t}$
\ENDFOR
\renewcommand{\algorithmicrequire}{\textbf{Output:}}
\renewcommand{\algorithmicensure}{\textbf{End}}
\REQUIRE $\{\mathbf{\Pi}^{t}, \mathbf{\Sigma}^{t}, N^{t}\}$, $\mathbf{U}^{t}$,$\mathbf{v}^{t}$, $\boldsymbol{\tau}^{t}$, $\{\mathbf{A}_i^{t}\}_{i=1}^d$, $\{\mathbf{b}_i^{t}\}_{i=1}^d$.
\end{algorithmic}
\end{algorithm}
Similar as (\ref{MoGLRMF1}), we can get the MAP problem:
\vspace{-2mm}
\begin{equation}
\small
\begin{split}
\mathcal{L}&(\mathbf{\Pi},\mathbf{\Sigma},\mathbf{v},\mathbf{U},\boldsymbol{\tau})=\\
&-\ln p(\mathbf{x}^t\circ\boldsymbol{\tau}|\mathbf{\Pi},\mathbf{\Sigma},\mathbf{v},\mathbf{U})+
\mathcal{R}_F^t(\mathbf{\Pi},\mathbf{\Sigma})+\mathcal{R}_B^t(\mathbf{U}).
\end{split}
\end{equation}
The key to solve this problem is to deduce the updating equation to $\boldsymbol{\tau}$. Since $\mathbf{x}\circ\boldsymbol{\tau}$ is a nonlinear geometric transform,  it's hard to directly optimize $\boldsymbol{\tau}$. So we consider to optimize the following reformulated problem:
\begin{equation}
\small\vspace{-0mm}
\begin{split}
\mathcal{L}(\mathbf{\Pi},\mathbf{\Sigma},\mathbf{v},\mathbf{U},\Delta\boldsymbol{\tau})=&
-\ln p(\mathbf{x}^t\circ\boldsymbol{\tau}+\mathbf{J}\Delta\boldsymbol{\tau}|\mathbf{\Pi},\mathbf{\Sigma},\mathbf{v},\mathbf{U})\\
&+
\mathcal{R}_F^t(\mathbf{\Pi},\mathbf{\Sigma})+\mathcal{R}_B^t(\mathbf{U}),
\end{split}\label{TMoGMFObjective}
\end{equation}
where $\mathbf{J}$ is the Jacobian of $\mathbf{x}$ with respect to $\boldsymbol{\tau}$. After we get $\Delta\boldsymbol{\tau}$, we add it into
 $\boldsymbol{\tau}$ to update transformation. This method iteratively approximates the original nonlinear transformation with a locally linear approximation~\cite{peng2012rasl,he2014iterative}.
Like OMoGMF, we also use online-EM algorithm to solve the problem. The updating rule of MoG parameters can use Eq. (\ref{OMS1}) and (\ref{OMS2}) by changing  $\mathbf{x}$ into $\mathbf{x}\circ\boldsymbol{\tau}+\mathbf{J}\Delta\boldsymbol{\tau}$.
And for $\mathbf{v}$ and $\Delta\boldsymbol{\tau}$, we need to solve the following problem:\vspace{-1mm}
\begin{equation}\small
\begin{aligned}
&\{\mathbf{v},\Delta\boldsymbol{\tau}\}=arg\min_{\mathbf{v},\boldsymbol{\tau}}||\mathbf{w}\odot(\mathbf{x}\circ\boldsymbol{\tau}+\mathbf{J}\Delta\boldsymbol{\tau}-\mathbf{U}\mathbf{v})||_{F}^2.\\
\end{aligned}
\end{equation}

\begin{table*}\scriptsize
\caption{Model comparison of typical transformed background subtraction methods }
\newcommand{\tabincell}[2]{\begin{tabular}{@{}#1@{}}#2\end{tabular}}  %
\centering\vspace{-3mm}
\begin{tabular}{c c c c c}
\shline
Method&\tabincell{c}{Foreground/Background\\Decomposition }&Objective Function&Constraint/Basic Assumption&Implementation Scheme\\
\hline
\shline
RASL~\cite{peng2012rasl} &$\boldsymbol{\tau}\circ\mathbf{X}=\mathbf{L}+\mathbf{S}$
&$\min_{\mathbf{L},\mathbf{S},\boldsymbol{\tau}}||\mathbf{L}||_*+\lambda||\mathbf{S}||_1$
&No&Offline\\
Ebadi et al.~\cite{ebadi2015efficient}&$\boldsymbol{\tau}\circ\mathbf{X}=\mathbf{L}+\mathbf{S}+\mathbf{E}$
&$\min_{\mathbf{L},\mathbf{S},\boldsymbol{\tau}}||\mathbf{E}||_F^2+\lambda||\mathbf{S}||_1$
&$rank(\mathbf{L})\leq k$&Offline\\
\hline
incPCP\_TI~\cite{rodriguez2015translational} &$\mathbf{x}^t=\mathbf{h}*(\boldsymbol{\alpha}\circ\mathbf{l})+\mathbf{s}+\mathbf{e}$&
$\min_{\mathbf{l},\mathbf{s},\mathbf{h},\boldsymbol{\alpha}}||\mathbf{e}||_F^2+\lambda_1||\mathbf{s}||_1+\lambda_2||\mathbf{h}||_1$
& $rank(\mathbf{L}^t)\leq k$
&Online\\
t-GRASTA~\cite{he2014iterative}&$\boldsymbol{\tau}\circ\mathbf{x}^t=\mathbf{U}\mathbf{v}+\mathbf{s}$&
$\min_{\mathbf{v}}||\mathbf{s}||_1$
&$\mathbf{U}^T\mathbf{U}=\mathbf{I}$&Online: Heuristically update $\mathbf{U}$\\
\textbf{t-OMoGMF} &$\boldsymbol{\tau}\circ\mathbf{x}^t=\mathbf{U}\mathbf{v}+\mathbf{e}$&
 \tabincell{c}{$\min_{\mathbf{\Pi},\mathbf{\Sigma},\mathbf{v},\mathbf{U},\boldsymbol{\tau}}
-\ln p(\boldsymbol{\tau}\circ\mathbf{x}^t|\mathbf{\Pi},\mathbf{\Sigma},\mathbf{v},\mathbf{U})$\\$+
\mathcal{R}_F^t(\mathbf{\Pi},\mathbf{\Sigma})+\mathcal{R}_B^t(\mathbf{U})$\\ }
&$e_{i}\sim \sum_{k=1}^K \pi_k\mathcal{N}(e|0,\sigma_k^2)$&Online: Optimize $\mathbf{U}$
\\
\shline
\end{tabular}\vspace{-3mm}
\label{table:tmodel}
\end{table*}

\vspace{-2mm}
\noindent By reformulating this weighted least square problem as:
$$\small{\{\mathbf{v},\Delta\boldsymbol{\tau}\}=arg\min_{\mathbf{v},
\boldsymbol{\tau}}||\mathbf{w}\odot(\mathbf{x}\circ\boldsymbol{\tau}
-\begin{bmatrix} \mathbf{U},-\mathbf{J} \end{bmatrix}\begin{bmatrix} \mathbf{v}\\\Delta\boldsymbol{\tau} \end{bmatrix})||_{F}^2,}$$
we can directly get its closed-form solution as:
\begin{equation}\small
\begin{aligned}
\begin{bmatrix} \mathbf{v}\\\Delta\boldsymbol{\tau} \end{bmatrix}=({\mathbf{T}}^Tdiag(\mathbf{w})^2
\mathbf{T})^{-1}{\mathbf{T}}^Tdiag(\mathbf{w})^2(\mathbf{x}\circ\boldsymbol{\tau}), \label{UpdateTau}
\end{aligned}
\end{equation}
where $\mathbf{T}=\begin{bmatrix} \mathbf{U},-\mathbf{J} \end{bmatrix}$.
Through iteratively updating all involved parameters $\mathbf{\Pi},\mathbf{\Sigma},\mathbf{v},\Delta\boldsymbol{\tau}$ as aforementioned, the objective function (\ref{TMoGMFObjective}) is monotonically increasing. After get the final outputs of these parameters, we can use similar strategy introduced in Sec. 3.3.2 to update the subspace $\mathbf{U}$.

The above transformed-OMoGMF (t-OMoGMF) algorithm is summarized in $\mathbf{Algorithm}$ \ref{alg2}. To better illustrate the function of the transformation operator $\boldsymbol{\tau}$, we show in Fig. \ref{Fig3} the separated foregrounds from the low-rank subspace obtained by $\mathbf{Algorithm}$ \ref{alg2} during iterations in a frame along gradually transformed video sequence. It is seen that the frame can be automatically adjusted to be well-aligned by the gradually rectified transform operator $\boldsymbol{\tau}$.
We list some typical transformed methods for background subtraction in Table \ref{table:tmodel} for easy comparison of their different properties.

\begin{table*}\small
\caption{F-measure(\%) comparison of all competing methods on all video sequences in Li dataset. Each value is averaged over all foreground-annotated frames in the corresponding video. The most right column lists the average performance of each competing method over all video sequences. The best result in each video sequence is highlighted in bold and the second best is in italic. The methods with bold titles denote the online methods.
}
\centering\vspace{-3mm}
\begin{tabular}{c c c c c c c c c c c}
\shline
\multirow{2}{*}{Methods}&\multicolumn{10}{c}{data}\\
\cline{2-11}&$airp.$ &$boot.$ &$shop.$ & $lobb.$ & $esca.$ & $curt.$ &$camp.$ & $wate.$&$foun.$& Average\\
\shline
RPCA~\cite{candes2011robust}&71.11&\textbf{67.67}&\textbf{72.79}&78.12&64.09&81.65&44.56&65.56&72.39&68.66\\
GODEC~\cite{zhou2011godec}&62.69&58.39&70.71&73.29&57.42&59.84&43.71&48.79&66.01&60.09\\
RegL1~\cite{zheng2012practical}&65.63&62.46&71.97&75.27&60.95&62.69&44.42&57.86&73.17&63.82\\
PRMF~\cite{wang2012probabilistic}&65.87&62.29&71.99&75.32&60.20&65.17&44.04&61.95&72.98&64.42\\
\emph{\textbf{OPRMF}}~\cite{wang2012probabilistic}&66.17&61.82&71.95&73.99&60.12&70.86&42.89&61.89&71.80&64.61\\
\emph{\textbf{GRASTA}}~\cite{he2012incremental}&61.87&58.07&71.47&60.98&57.26&68.20&44.53&75.88&69.23&63.05\\
\emph{\textbf{incPCP}}~\cite{rodriguez2016incremental}&59.84&62.47&71.28&75.83&45.59&61.10&44.55&74.94&70.49&62.90\\
\emph{\textbf{PracReProCS}}~\cite{guo2014online}&70.01&\emph{63.71}&71.61&61.89&56.08&77.74&42.28&\emph{87.53}&62.76&65.96\\
\emph{\textbf{OMoGMF}}&\emph{74.08}&59.87&71.80&78.01&61.42&86.08&44.48&87.34&71.78&70.54\\
\hline
DECOLOR~\cite{zhou2013moving}&63.98&59.97&65.37&68.93&\emph{75.93}&\emph{89.56}&\textbf{77.14}&64.03&\textbf{86.76}&\emph{72.41}\\
\emph{\textbf{GOSUS}}~\cite{xu2013gosus}&65.80&61.95&72.12&\emph{80.97}&\textbf{86.27}&68.26&51.30&84.37&73.15&71.35\\
\emph{\textbf{OMoGMF+TV}}&\textbf{77.20}&61.17&\emph{72.43}&\textbf{83.47}&66.37&\textbf{92.54}&\emph{65.88}&\textbf{93.14}&\emph{82.53}&\textbf{77.19}\\
\shline
\end{tabular}
\label{table:Fs}
\end{table*}

As aforementioned, we need a warm-start stage on a mini-batch of video frames for subspace initialization. We thus also need to pre-align these starting frames to facilitate a good starting point for the following online implementation. To this aim we extend $\mathbf{Algorithm}$ \ref{alg2} as a frame-recursive-updating version (we call it as iterative t-OMoGMF or it-OMoGMF) as follows: first easily set the mean or the median of all starting frames as an initial subspace, and then iterate the following two steps: run $\mathbf{Algorithm}$ \ref{alg2} throughout all these frames and then return to the first frame and rerun the algorithm. During this process, the separated foreground and background are expected to be more and more accurate due to the gradually ameliorated transformation operators on all frames.
Throughout all our experiments, no more than $4$ iterations of the algorithm can follow a satisfactory initialization.
Note that this algorithm can also be separately understood as an offline subspace learning method capable of simultaneously implement transformation learning and background subtraction. It can thus be extended to more applications, like image alignments and video stabilization, which will be discussed in Sec. 5.1.

Note that we can easily design an adaptive algorithm to determine whether to activate t-OMoGMF from the implementation of OMoGMF by adding a judging condition during iterations of OMoGMF. Two strategies can be considered to set this condition: to judge whether the sparsity of the foreground temporally extracted from the video is of an evident increase, and to use t-OMoGMF for some equidistant frames to detect whether the calculated transformation operator is an approximate identity mapping. Both strategies can be easily formulated. Such amelioration tends to facilitate the proposed strategy to better adapt real scenes.

\begin{figure}[!]
\centering
\subfigure[Original frames]{
\label{Fig.sub.1}
\includegraphics[width=0.23\textwidth]{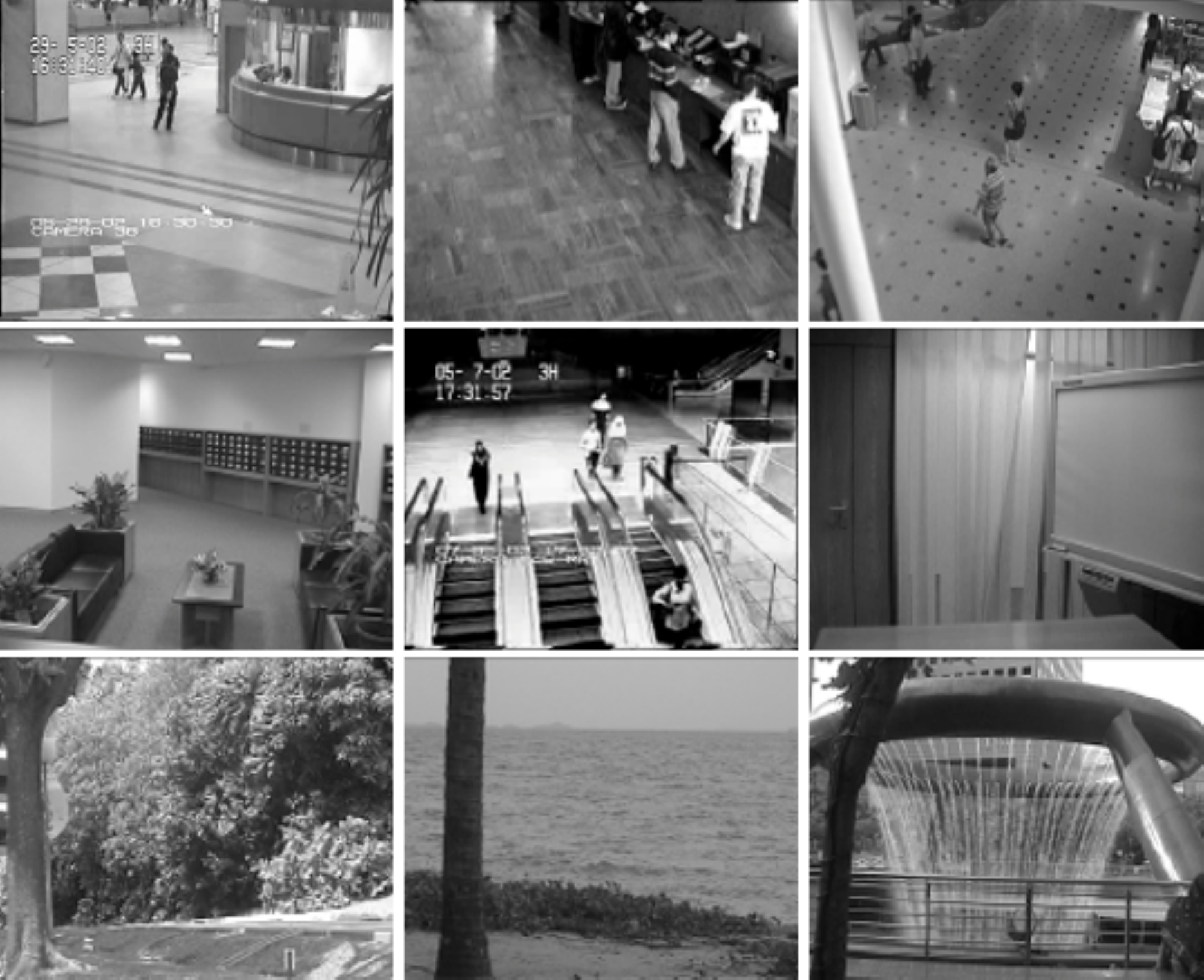}}
\subfigure[Transformed frames]{
\label{Fig.sub.2}
\includegraphics[width=0.23\textwidth]{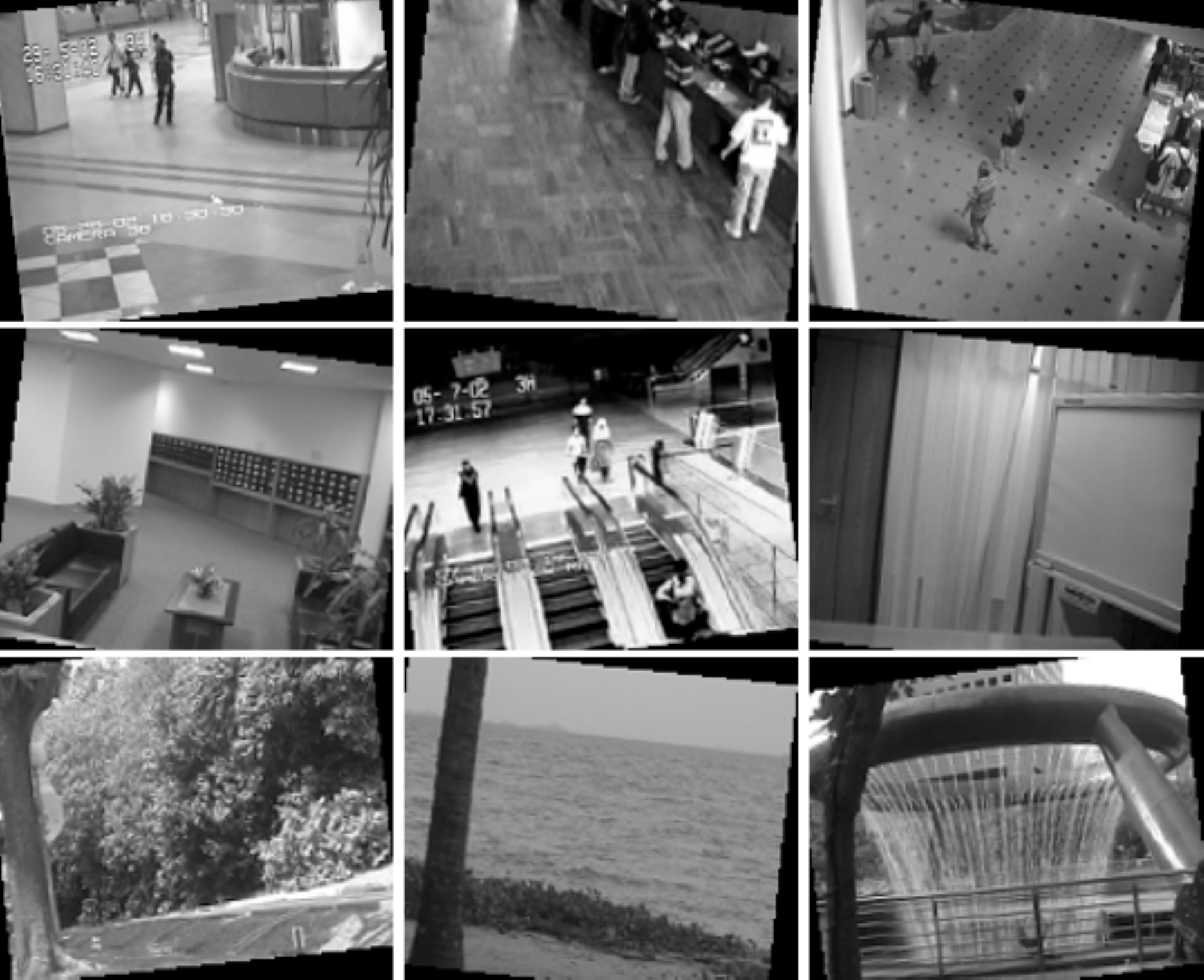}}
\vspace{-4mm}
\caption{Typical frames in $9$ video sequences of the original and transformed \emph{Li} dataset.}
\label{Lidata}
\vspace{-3mm}
\end{figure}

\section{Experiments}
In this section we depict the performance of the proposed methods on videos with static and dynamic backgrounds, respectively. All experiments were implemented on a personal computer with i7 CPU and 32G RAM.

\subsection{On videos without camera jitter}
\textbf{Dataset.} In this experiments we employ the Li dataset\footnote{http://perception.i2r.a-star.edu.sg/bk\_model/bk\_index.html}, including $9$ video sequences, each pictured under a fixed survivance camera to certain scene. Typical frames in these videos are depicted in Fig. \ref{Lidata}(a). These video sequences range over a wide range of background cases, like static background (e.g., \emph{airport}, \emph{bootstrap}, \emph{shoppingmall}), illumination changes (e.g., \emph{lobby}), dynamic background
indoors (e.g., \emph{escalator}, \emph{curtain}) and dynamic background outdoors (e.g., \emph{campus}, \emph{watersurface}, \emph{fountain})\footnote{We use the first three/four letters of each sequence name as its abbreviation in later experiments.}. For each video, we choose 200 frames for training. Each sequence contains multiple frames with pre-annotated groundtruth foregrounds and they are also added into the training data for evaluating the accuracy of foreground region by each competing method.

\textbf{Comparison methods.} We adopted a series of typical offline and online state-of-the-art background subtraction methods for experiment comparison. The utilized offline methods include: PRCA\footnote{http://perception.csl.illinois.edu/matrix-rank/home.html},
GODEC\footnote{https://tianyizhou.wordpress.com/},
RegL1\footnote{https://sites.google.com/site/yinqiangzheng/},
PRMF\footnote{http://winsty.net/prmf.html\label{footnote1}},
DECOLOR\footnote{http://bioinformatics.ust.hk/decolor/decolor.html},
and online methods include:
OPRMF$^{\ref{footnote1}}$,
GOSUS\footnote{http://pages.cs.wisc.edu/~jiaxu/projects/gosus/},
GRASTA\footnote{http://sites.google.com/site/hejunzz/grasta},
incPCP\footnote{https://sites.google.com/a/istec.net/prodrig/Home/en/pubs/incpcp},
PracReProCS\footnote{http://www.ece.iastate.edu/\%7Ehanguo/PracReProCS.html}.
Beyond other competing methods, DECOLOR, GOSUS and OMoGMF+TV specifically consider the spatial smoothness over the foreground region, making them always get higher accuracy in capturing foreground objects.
\begin{figure*}[thb!]
\centering
\subfigure{
\includegraphics[width=0.98\textwidth]{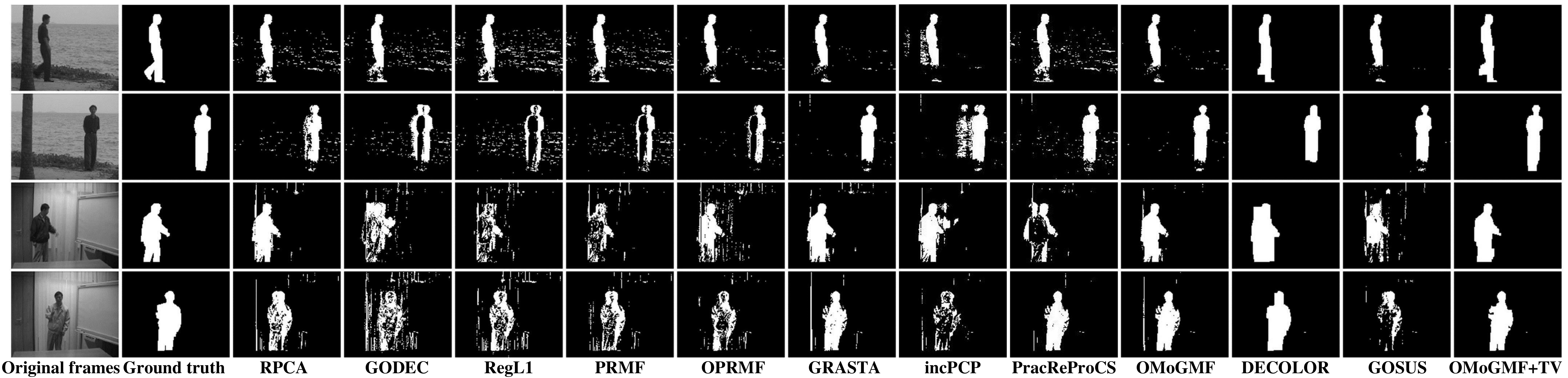}}\vspace{-4mm}
\caption{From left to right: typical frames from $curtain$ and $watersurface$ sequences, groundtruth foreground objects, foregrounds detected by all competing methods.}
\label{TypicalBSEffect}\vspace{-3mm}
\end{figure*}

\begin{table}\small
\caption{FPS of online competing methods on three videos, each with 200 frames while with different frame size, in Li dataset.
}
\centering\vspace{-3mm}
\begin{tabular}{c c c c}
\shline
Video&$esca.$&$airp.$&$shop.$\\
\hline
Frame Size&130$\times$ 160&144$\times$ 176&256$\times$  320\\
\shline
OPRMF~\cite{wang2012probabilistic}&0.5&0.4&0.1\\
PracReProCS~\cite{guo2014online}&1.5 &1.2 &0.2\\
GOSUS~\cite{xu2013gosus}&3.8&2.7&0.6\\
\textbf{OMoGMF+TV} &18.5&14.8&3.5\\
\textbf{OMoGMF} &99.6&63.0&5.2\\
GRASTA~\cite{he2012incremental}&\emph{166.9}&\emph{123.9}&\emph{28.7}\\
incPCP~\cite{rodriguez2016incremental}&\textbf{274.5}&\textbf{220.8}&\textbf{85.2}\\
\hline
GRASTA\textbf{\&}1\%{SS} &303.2&246.7&65.5\\
\textbf{OMoGMF}\textbf{\&}1\%{SS}  &\textbf{332.0}&\textbf{263.6}&\textbf{104.7}\\
\shline
\end{tabular}\vspace{-3mm}
\label{table:FPS}
\end{table}
\textbf{Experiments setup.}
For RPCA, RegL1, PRMF, OPRMF, DECOLOR, GRASTA,incPCP  and PracReProCS, we use the default parameter settings in the original codes. For GODEC, we set the sparse parameter by using the result of RPCA. For GOSUS, we set $\lambda$ using cross-validation, and set others by default settings. For all low-rank methods, we empirically set a proper rank for each video, and this rank parameter is set the same for all competing methods in each video. For all online competing methods, we randomly choose some frames and use robust batch method and PCA to initialize the subspace. We use GMM on the residuals to initialize the MoG parameters for OMoGMF, and the number of MoG components is set as $3$ throughout all our experiments. The foreground obtained by the proposed methods are taken as the MoG noise component with the largest variance. More details on parameter setting in our experiments are introduced in SM.
\begin{table*}\small
\caption{F-measure(\%) and FPS of OMoGMF and GRASTA under different sub-sampling rates on $3$ videos, each with 1000 frames and different frame size, in Li dataset. The average is computed on the results of both methods on all $9$ videos of Li dataset. Detailed results are reported in SM.}
\centering\vspace{-3mm}
\begin{tabular}{ c c c c c c c c c c c c c }
\shline
\multicolumn{3}{c}{Sub-Sampling rate}  & \multicolumn{2}{c}{1\%}  & \multicolumn{2}{c}{10\%}&\multicolumn{2}{c}{30\%}&
\multicolumn{2}{c}{50\%}& \multicolumn{2}{c}{100\%} \\
\hline
Dataset&frame size & method &F-M & FPS & F-M & FPS & F-M & FPS & F-M & FPS & F-M&FPS\\
\shline
\multirow{2}{*}{$esca.$}&\multirow{2}{*}{130$\times$ 160}&
OMoGMF & 61.01&  332.0 &   61.19 & 265.9   & 61.26  &172.3  &  61.28  &137.5  &  61.20  &  99.6\\
&&GRASTA & 45.81&  303.2 &   58.97 & 264.6   & 58.70  &212.8  &  58.21  &180.2  &  57.32  & 166.9\\
\hline
\multirow{2}{*}{$airp.$}&\multirow{2}{*}{144$\times$ 176}&
OMoGMF  &71.31&  263.6 &   72.30 &  181.6  &  72.38 & 115.7 &   72.42 &  91.7 &   72.41 &  63.0 \\
&&GRASTA &63.12&  246.7 &   64.03 &  209.6  &  63.25 & 167.6 &   62.56 & 141.3 &   61.94 & 123.9\\
\hline
\multirow{2}{*}{$shop.$}&\multirow{2}{*}{256$\times$  320}&
OMoGMF &  69.42 & 104.7 &   69.46 &  40.2  & 69.47   & 16.1  &  69.48  & 10.2  &  69.50  &  5.2\\
&&GRASTA &  65.35&   65.5 &   71.00 &  56.7   & 71.43  & 44.2  &  71.51  & 37.1  &  71.46  & 28.7\\
\hline
\multirow{2}{*}{$Average$}&\multirow{2}{*}{\_}&
OMoGMF  &\textbf{69.82} & \textbf{273.4} &  \textbf{70.30} &189.4 & \textbf{70.33} & 114.2 & \textbf{70.35} & 86.1 & \textbf{70.34} & 57.6\\
&&GRASTA &57.16 & 244.0 &  64.30& \textbf{213.9} & 65.16  &\textbf{171.1} & 64.58 &\textbf{145.7} & 63.07 & \textbf{128.7}\\
\shline
\end{tabular}
\label{table:t3}
\end{table*}

\textbf{Performance Evaluation.}
The F-measure is utilized as the quantitative metric for performance evaluation.
The F-measure is calculated as follows:\\
$$
\text{\emph{F-measure}}=2\times\frac{precision\cdot recall}{precision+recall},
$$
where $precision=\frac{|\mathbf{S}_{f}\bigcap \mathbf{S}_{gt}|}{|\mathbf{S}_{f}|}$ and $recall=\frac{|\mathbf{S}_{f}\bigcap \mathbf{S}_{gt}|}{|\mathbf{S}_{gt}|}$, $\mathbf{S}_{f}$ and $\mathbf{S}_{gt}$ denote the support sets of the foreground calculated from the method and the ground truth one, respectively. F-measure is close to the smaller one of $precision$ and $recall$. Besides, we choose the value of the threshold which can make $\text{\emph{F-measure}}$ largest for each competing method. The larger the $\text{\emph{F-measure}}$, the more accurate the foreground object is detected by the method.

\textbf{Foreground detection accuracy comparison.} Table \ref{table:Fs} lists the average F-measure over all foreground-annotated frames calculated from all competing methods on all video sequences in Li dataset. It is seen that in average OMoGMF has an evidently better performance than other competing methods without considering spatial smoothness of videos, and so does OMoGMF+TV than all. This superiority of the OMoGMF and OMoGMF+TV methods can also be observed from Fig. \ref{TypicalBSEffect}, which shows the detected foregrounds on typical frames of tested videos by all competing methods. It is easy to see that the detected foregrounds by the proposed methods are closer to the groundtruth ones, which conducts its larger F-measures in experiments.

The advantage of the proposed methods can be easily explained by their flexible noise fitting capability along time. Specifically, while other competing methods only assume a fixed noise distribution throughout a video sequence and separate one layer of foreground from the video, our methods can adapt a specific MoG noise for each video frame, and thus can more elaborately recover multiple layers of foreground knowledge. Especially, the extracted noise component with largest variance generally dominates the foreground object, while other noise components
characterize smaller variations accompanied besides the object. E.g., for $lobby$ and $campus$ sequences, except that the noise component with largest variance properly delivers the moving objects in the video, the other two components finely capture the shadow along the object + weak camera noise and
leaves shaking variance + weak camera noise, respectively, as clearly depicted in Fig. \ref{figure:f1}. Due to such ability of clearing away trivial noise variations, our methods then rationally attain
better foreground object extraction performance\footnote{More demos on depicting such noise fitting capability of the proposed methods are shown in http://gr.xjtu.edu.cn/web/dymeng/7.}.

\begin{figure}[!]
\centering
\subfigure[Video sequence with illumination change]{
\label{Fig.sub.1}
\includegraphics[width=0.38\textwidth]{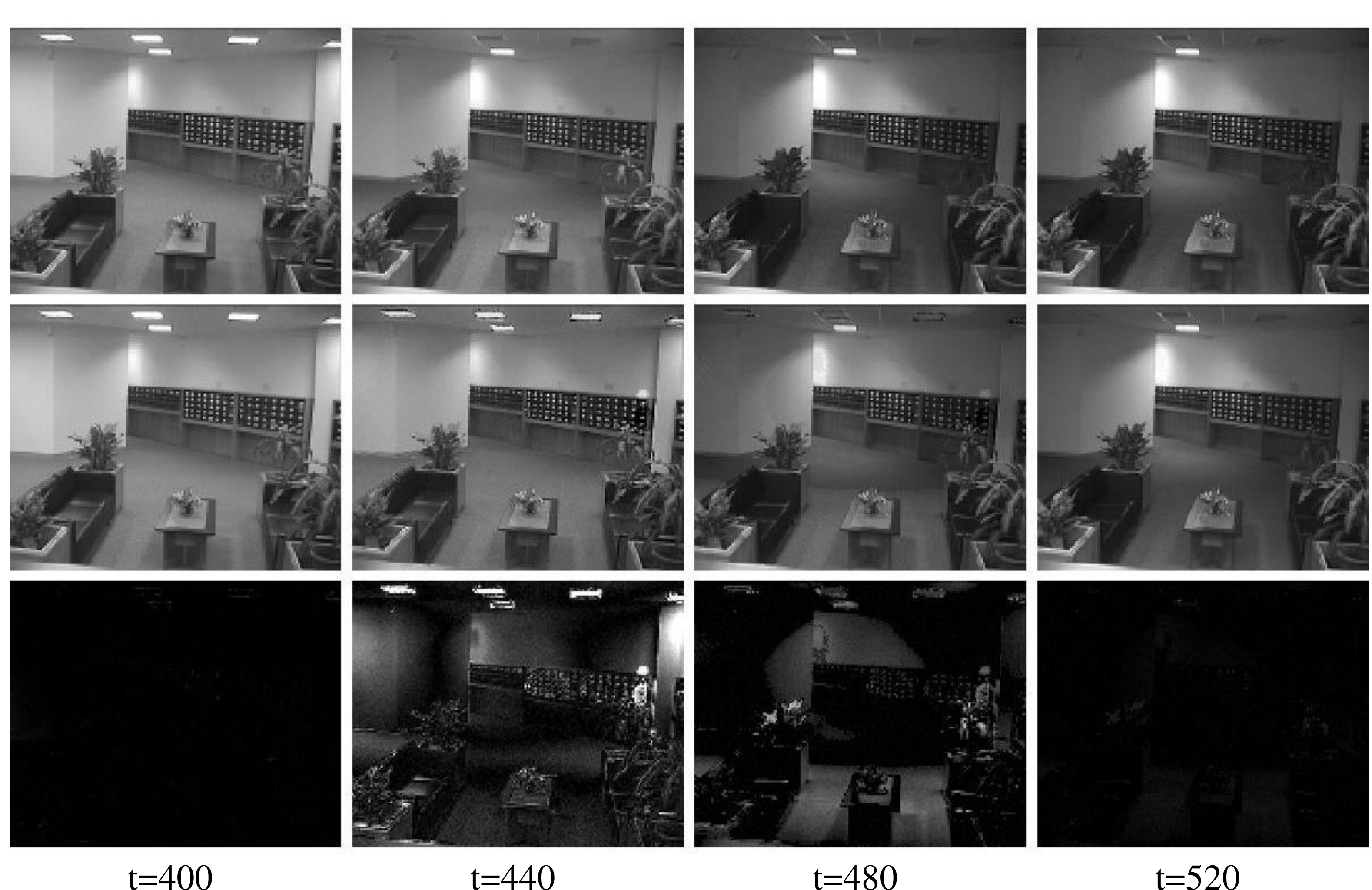}\vspace{-1mm}}
\subfigure[Sequence with missing entries]{
\label{Fig.sub.2}
\includegraphics[width=0.38\textwidth]{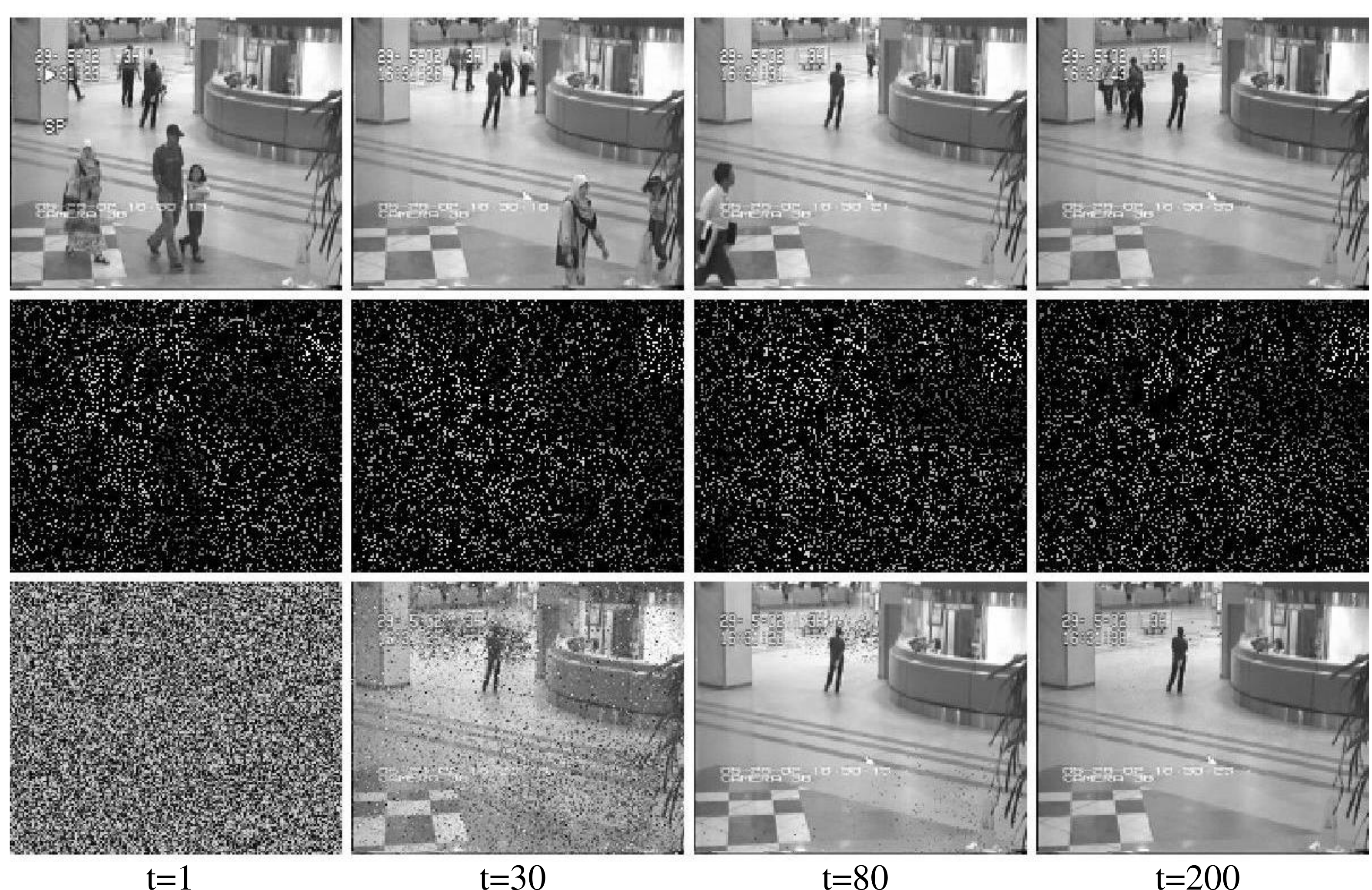}\vspace{-1mm}}\vspace{-3mm}
\caption{Performance of OMoGMF on video sequences with (a) sudden illumination change and (b) missing entries. From upper to lower: (a) Original video frames, extracted backgrounds and residuals; (b) Original video frames, 15\% subsampling frames, extracted backgrounds.
}
\label{figure:ChangeAndMissing}
\vspace{-2mm}
\end{figure}

\textbf{Effects on videos with background changes and missing entries.}
Beyond offline methods, which are relatively sensitive to background changes, the proposed online method can finely adapt sudden background changes. Fig. \ref{figure:ChangeAndMissing}(a) shows the background subtraction result by OMoGMF for frames in \emph{lobby} sequence with a illumination change. It is seen that OMoGMF can gradually adapt a proper background after multiple frames' self-ameliorating. Furthermore, OMoGMF can also be used to deal with videos with missing entries by adopting the sub-sampling technique. Fig. \ref{figure:ChangeAndMissing}(b) shows background subtraction result on \emph{airport} sequence, with 85\% of its entries missed under a random subspace initialization. It is easy to see that our online method can gradually recover the proper background.

\begin{figure}[!]
\centering\hspace{-2mm}
\subfigure{
\includegraphics[width=0.4\textwidth]{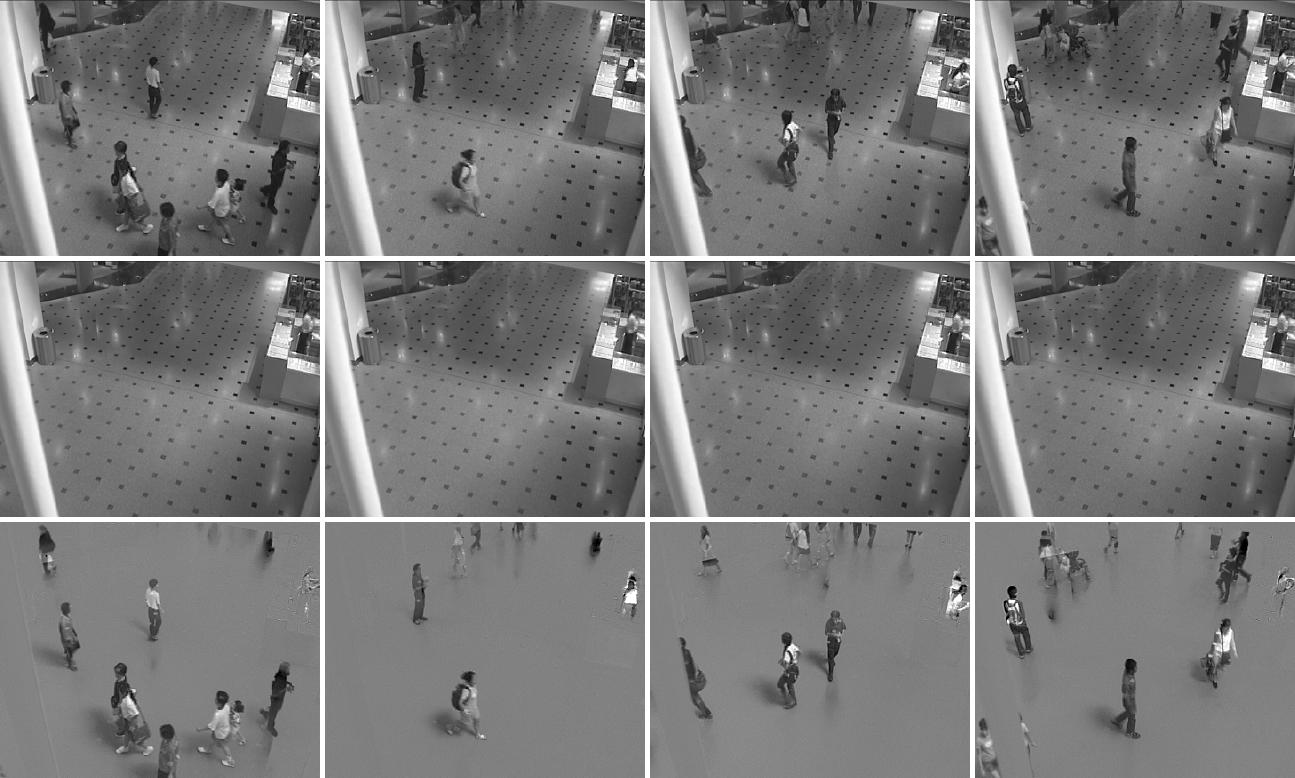}}\vspace{-3mm}
\caption{From upper to lower: original frames in the \emph{shoppingmall} video, backgrounds and residuals (brightening 0.5 gray scales for all pixels for better visualization) obtained by OMoGMF with 1\% sub-sampling rate.}
\label{figure:subsample}\vspace{-3mm}
\end{figure}

\textbf{Speed Evaluations.} In our experiments, just as substantiated by previous research, all offline competing methods are significantly slower than online methods, which is especially more evident for those videos with relatively more frames and larger sizes. So here we only compare the speed  of the online competing methods. The FPS (frames per second) is utilized as the comparison metric.
\begin{table*}\small
\caption{F-measure(\%) of different competing methods on Li dataset added with artificial transformations.}
\centering\vspace{-3mm}
\begin{tabular}{ c c c c c c c c c c c }
\shline
\multirow{2}{*}{Methods}&\multicolumn{10}{ c }{data}\\
\cline{2-11}&$airp.$ &$boot.$ &$shop.$ & $lobb.$ & $esca.$ & $curt.$ &$camp.$ & $wate.$&$foun.$& Average\\
\shline
RPCA~\cite{candes2011robust}&58.53 &59.10 & 53.67 & 19.48 &44.87& 62.04& 32.81&  51.40&  38.49&46.71\\
GRASTA~\cite{he2012incremental}&52.96 & 50.38 &41.38 & 9.19&32.85 & 58.54 & 24.23 & 44.22 &19.93&37.08\\
OMoGMF&51.97 & 49.35 & 39.71 & 9.36& 28.19 & 66.77&21.05& 74.35& 19.90&40.07\\
\hline
RASL~\cite{peng2012rasl}&68.94&\textbf{66.46} &\textbf{76.42}& \emph{61.17}& \textbf{72.53}& 69.91& \textbf{58.87}& 30.18& 71.17&63.96\\
t-GRASTA~\cite{he2014iterative}&\emph{69.23} &56.91 &46.57&56.35&64.26& \emph{83.01} &51.15& \emph{88.34} & \emph{72.89}&\emph{65.41}\\
\textbf{t-OMoGMF}& \textbf{75.21} &\emph{65.08} & \emph{74.92}  &  \textbf{65.57} & \emph{66.16} & \textbf{89.62}  & \emph{53.65}  & \textbf{89.79} & \textbf{79.41}&\textbf{73.27}\\
\shline
\end{tabular}
\label{table:t4}
\end{table*}
Table \ref{table:FPS} shows the FPS of all online competing methods on three videos with different frame sizes. Though the FPS of OMoGMF is not the highest,  the OMoGMF can achieve the real-time processing in most case. Sub-sampling technique can be easily used  into  GRASTA and OMoGMF to further accelerate their computation.
We also show the FPS of GRASTA and OMoGMF with 1\% subsampling rate in this table.
Under low subsampling rate,  OMoGMF can attain more than 200 FPS on average for executing these videos while also keep a high F-measure so that it can meet the real-time requirement in practice. To better clarify this point, Table \ref{table:t3} shows the performance of OMoGMF and GRASTA under different sub-sampling rates on $3$ videos with different sizes, each with 1000 frames. It can be seen that as the sub-sampling rate decreasing, OMoGMF is gradually more accurate and faster than GRASTA in all videos. Specifically, in $5$ of $9$ videos, GRASTA has an evident drop in F-measure\footnote{On multiple videos, the GRASTA method has a surprising increase on F-measure under 1\% sub-sampling rate than the method on the original videos. This might be due to the fact that the sub-sampling technique helps alleviate the negative influence of the complex noise configurations, especially in the videos with dynamic backgrounds. }. Comparatively, even under 1\% sub-sampling rate, the F-measure obtained by OMoGMF is only very slightly decreased than that on the original videos. As shown in Fig. \ref{figure:subsample}, even under such a low sub-sampling rate, OMoGMF can still finely capture foreground and background of original videos.


\begin{table}\small
\caption{F-measure(\%) comparison of transformed methods on real data with
camera jitter
}
\centering\vspace{-3mm}
\begin{tabular}{c c c c c c }
\shline
\multirow{2}{*}{Methods}&\multicolumn{5}{c}{data}\\
\cline{2-6}&$badm.$ &$boul.$ &$side.$ & $traf.$ & Average\\
\shline
RASL ~\cite{peng2012rasl}   &\textbf{72.86}  &  \textbf{73.62}  &  47.16 &   \emph{67.77}&   65.35\\
t-GRASTA~\cite{he2014iterative}&68.99  &  \emph{72.62}  &  \emph{56.20} &   67.28&   \emph{66.27}\\
t-OMoGMF&\emph{69.97}  &  69.78  &  \textbf{59.20} &   \textbf{83.44} &   \textbf{70.60}\\
\shline
\end{tabular}
\label{table:t-realdata}
\end{table}

\subsection{On videos with camera jitter}
We then test the t-OMoGMF method on videos with camera jitter. The comparison methods include state-of-the-art subspace alignment methods: RASL\footnote{http://perception.csl.uiuc.edu/matrix-rank/rasl.html} and
t-GRASTA\footnote{https://sites.google.com/site/hejunzz/t-grasta}.

\begin{figure}[!]
\subfigure{
\includegraphics[width=0.47\textwidth]{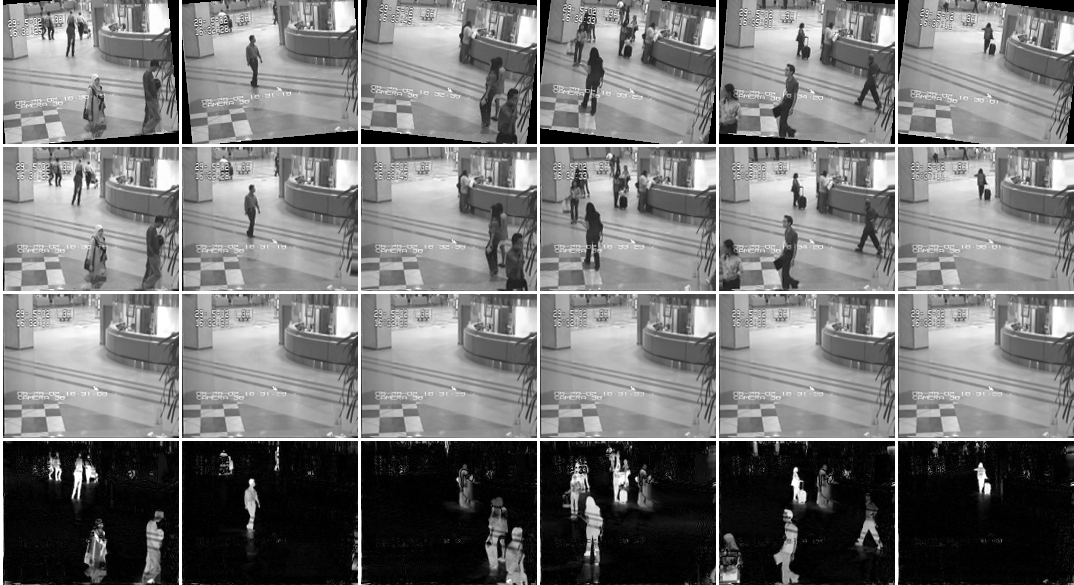}}\vspace{-3mm}
\caption{From upper to lower: original frames in \emph{airport} video; aligned frames, backgrounds and foregrounds obtained by t-OMoGMF.
 }
\label{figure:Hallsubsample}\vspace{-4mm}
\end{figure}

\textbf{Synthetic data}: As \cite{he2014iterative}, we design $9$ synthetic jitter video sequences from Li dataset with background transformations. $400$ adjacent frames are selected from each video, and each frame is randomly rotated with an angle in [$-5^\circ,5^\circ$] and translated in both axes with a range in [-5,5], respectively. Typical frames of all videos so generated are shown in Fig. \ref{Lidata}(b). We randomly choose 30 frames for each video for subspace initialization. The rank is set as $3$ for all competing methods.
Table \ref{table:t4} shows the F-measure obtained by t-OMoGMF, RASL and t-GRASTA in all test videos. Besides, the results of PRCA, GRASTA and OMoGMF, which do not consider frame transformations, are also compared to show the function of the involvement of this transformation operator. The superiority of t-OMoGMF can be easily observed: It performs the best/the second best in 5/4 out of 9 videos, and on average it performs more than 10\% better than other methods.
For better observation, Fig. \ref{figure:Hallsubsample} shows the result of t-OMoGMF on multiple frames of transformed $airport$ sequence. It is easy to see that
t-OMoGMF can well align video frames, which naturally leads to its better foreground/background separation. About computational speed, our method can run more than 5 times faster than RASL and slightly faster than t-GRASTA on average.

\begin{figure}[!]
\centering
\subfigure{
\includegraphics[width=0.46\textwidth]{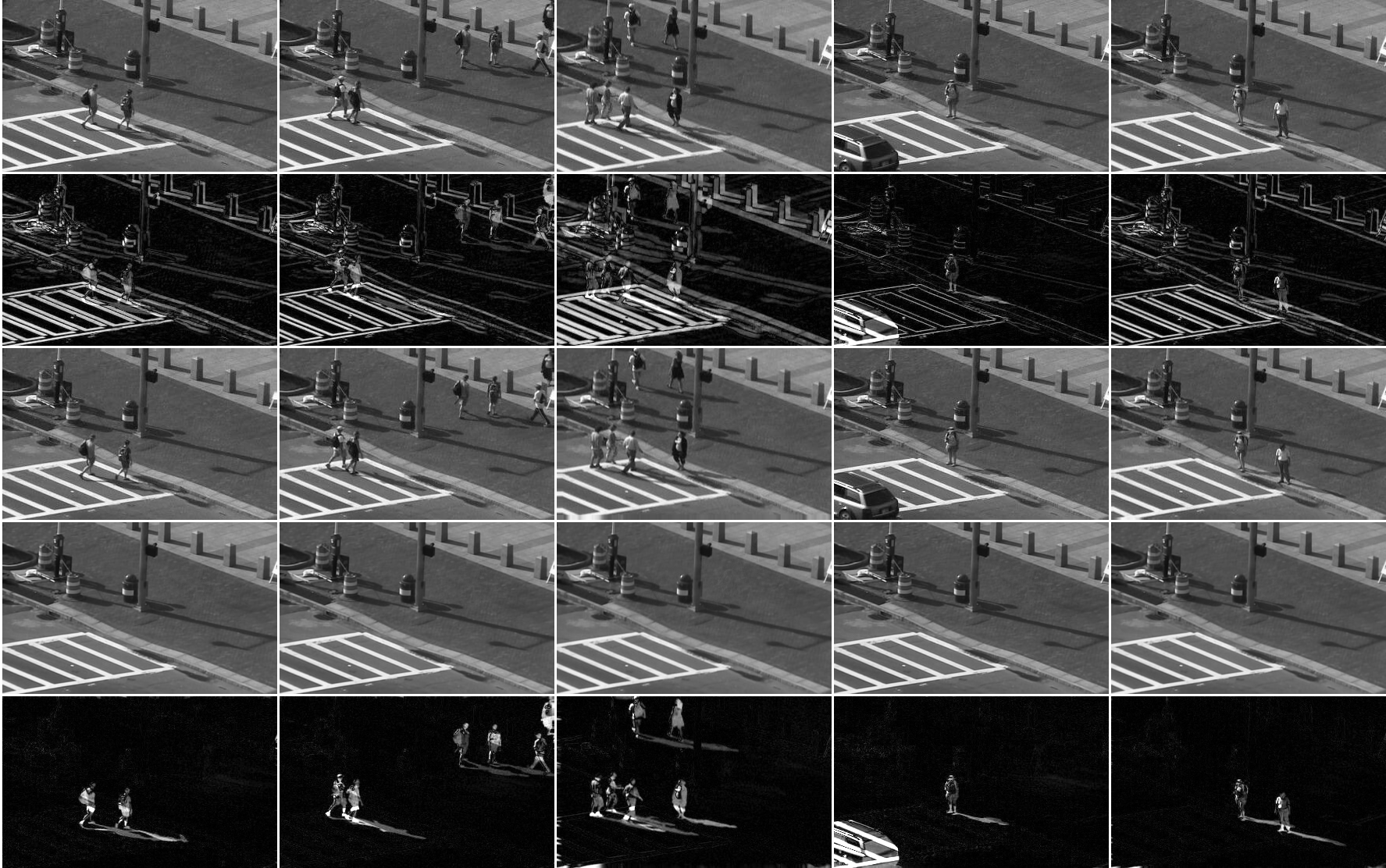}}\vspace{-3mm}
\caption{From upper to lower: original frames in \emph{sidewalk} video; residuals on unaligned frames; frames aligned by t-OMoGMF; backgrounds obtained by t-OMoGMF; foregrounds detected  by t-OMoGMF.}
\label{figure:sidewalk}
\end{figure}

\textbf{Real data}: In this experiment, we use a set of real-world videos for assessing the performance of the proposed t-OMoGMF method. The dataset contains 4 video sequences\footnote{http://www.changedetection.net}: \emph{badminton}, \emph{boulevard}, \emph{sidewalk} and \emph{traffic}, all with evident camera jitters. As set in synthetic experiments, we set subspace rank as 3 for all competing methods, and randomly choose $30$ frames from each video to train the initial subspace and model parameters. In all our experiments, t-OMoGMF can properly align all video frames, and simultaneously appropriately separate background and foreground from the jitter videos\footnote{Please see more demos in http://gr.xjtu.edu.cn/web/dymeng/7.}. Some typical frames are shown in Fig. \ref{figure:sidewalk}.
Furthermore, Table \ref{table:t-realdata} quantitatively compares the performance of all competing methods in terms of the F-measure on these real videos, and the superiority of the proposed t-OMoGMF can still be observed on average. Besides, while RASL and GRASTA may also detect foreground objects from videos, our method can always more clearly recover the background and clean up various unexpected noise variations from the video, as clearly depicted in Fig. \ref{Fig2}.

\section{Conclusion and discussion}

\subsection{Extension to other tasks}
As aforementioned in Section 3.6, the proposed it-OMoGMF method can be easily extended to image alignment and video stabilization tasks. Here we provide some preliminary tests to illuminate this point.

\begin{figure}
\centering
\subfigure{
\includegraphics[width=0.45\textwidth]{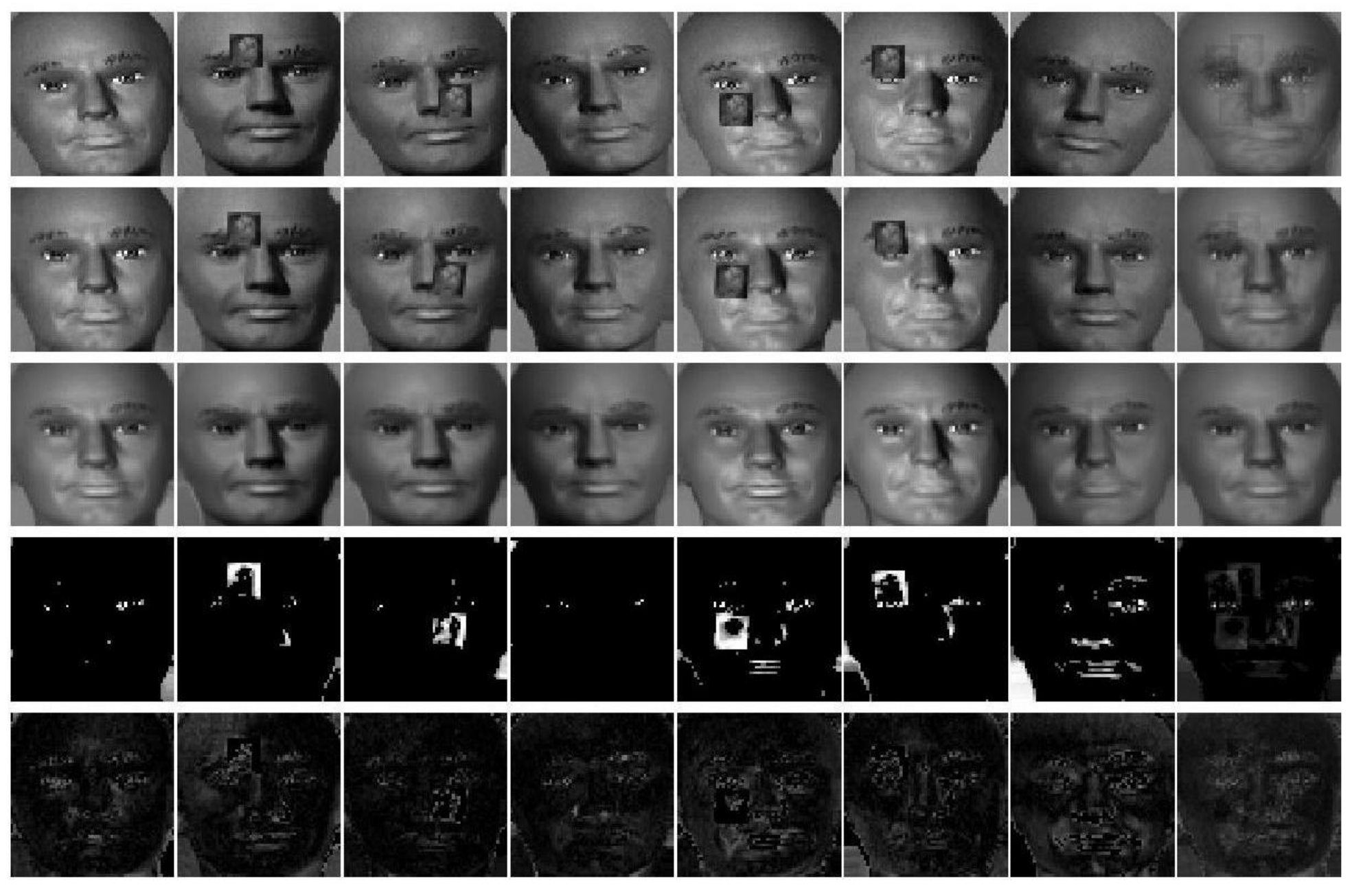}}
\vspace{-5mm}
\caption{From upper to lower: original images in ``dummy" set; images aligned by it-OMoGMF; images recovered by it-OMoGMF; occlusions and small variations detected by it-OMoGMF. The last column is averaged over all images. }
\label{figure:faceimage}
\vspace{-3mm}
\end{figure}

\begin{figure}
\hspace{-2mm}
\subfigure{
\hspace{-1mm}\includegraphics[width=0.5\textwidth]{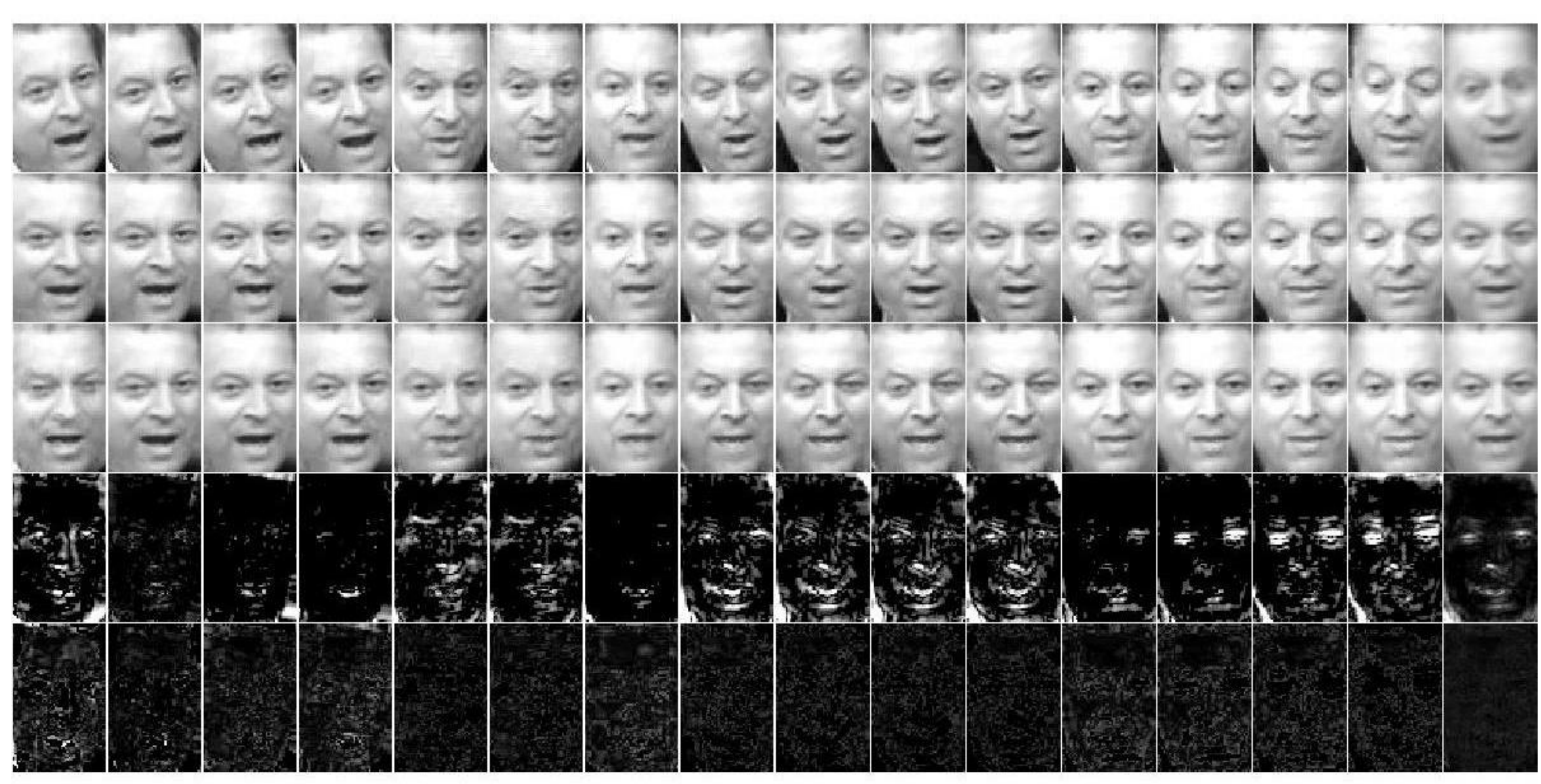}}
\vspace{-7mm}
\caption{From upper to lower: original frames from ``Gore" video; frames aligned by it-OMoGMF; frames recovered by it-OMoGMF; expression changes and small variations  detected by it-OMoGMF. The last column is averaged over all frames.}
\label{figure:facevideo}
\vspace{-3mm}
\end{figure}

Two datasets were adopted in this experiment: one is the ``dummy" imageset~\cite{peng2012rasl}, a synthetic face dataset containing $100$ misaligned faces with block occlusions and illumination variations; the other is the ``Gore" video sequence~\cite{peng2012rasl}, a 140-frame video of Al Gore talking, obtained by applying
a face detector to each frame independently. Due to inherent imprecision of the detector, there is significant jitters from frame to frame~\cite{peng2012rasl}. As previous state-of-the-art methods implemented on these datasets, the proposed it-OMoGMF method can also finely align all of the images/frames in two datasets, as clearly demonstrated in Fig. \ref{figure:faceimage} and \ref{figure:facevideo}, respectively. Yet beyond other methods, it-OMoGMF can more elaborately separate multiple layers of noises from data, which always have certain physical meanings. E.g., the noise components with largest variance in two datasets clearly represent the occlusion part for ``dummy" images and expression variations for "Gore" video, respectively.

\subsection{Concluding remarks}

In this paper, we have proposed a new online subspace learning method aiming to make background subtraction available in practical videos both in speed and accuracy. On one hand, the computational speed of the new method reaches the real-time requirement for video processing (more than 25 FPS), and on the other hand, the method can adaptively fit real-time dynamic variations in both foreground and background of videos. In particular, through specifically learning a foreground distribution and a background subspace regularized by the previously learned knowledge for each video frame, the method can properly deliver variations of video foreground and background along the video sequence. Through further involving the variables to encode affine transformation operators on each video frame, the method can further adapt the background transformations, like rotations, transformations, scalings and distortions, generally existed in practical videos. Furthermore, by virtue of the sub-sampling and TV-norm regularization technique amelioration, the efficiency and accuracy of the proposed method can be further improved. The superiority of the proposed method has been extensively substantiated, as compared with other state-of-the-art online and offline methods along this research line, by experiments on synthetic and real-world videos.

\subsection{Future work}
In our future investigations, we will try to embed the sub-sampling technique into t-OMoGMF and it-OMoGMF to further improve their speed and apply our methods to larger real-time video streaming. Also, more video processing tasks are worthy to be intrinsically integrated with the proposed methods, especially to make full use of the real-timely detected foreground objects by the proposed methods. Recently, there are multiple literatures discussing the parameter selection issue, like the subspace rank~\cite{zhao2014robust} and the number of noise components~\cite{cao2015moep}, in subspace learning. We'll also consider to combine these techniques into our method to make its parameter selection more proper and automatic. Besides, we will try to extend our method to other computer vision and image processing applications.

\end{document}